\documentclass{article}

\usepackage{cite}
\usepackage{amsmath,amssymb,amsfonts}
\usepackage{algorithmic}
\usepackage{graphicx}
\usepackage{textcomp}
\def\BibTeX{{\rm B\kern-.05em{\sc i\kern-.025em b}\kern-.08em
    T\kern-.1667em\lower.7ex\hbox{E}\kern-.125emX}}
\usepackage[a4paper, total={6in, 8in}]{geometry}
    
\usepackage[nolist,nohyperlinks]{acronym}
\usepackage{enumitem}
\usepackage{url}
\usepackage{hyperref}
\usepackage{tikz}

\definecolor{lime}{HTML}{A6CE39}
\DeclareRobustCommand{\orcidicon}{
	\begin{tikzpicture}
	\draw[lime, fill=lime] (0,0) 
	circle [radius=0.16] 
	node[white] {{\fontfamily{qag}\selectfont \tiny ID}};
	\draw[white, fill=white] (-0.0625,0.095) 
	circle [radius=0.007];
	\end{tikzpicture}
	\hspace{-2mm}
}
\foreach \x in {A, ..., Z}{\expandafter\xdef\csname orcid\x\endcsname{\noexpand\href{https://orcid.org/\csname orcidauthor\x\endcsname}
			{\noexpand\orcidicon}}
}

\newlist{questions}{enumerate}{1}
\setlist[questions]{align=left,font=\itshape}
\setlist[questions,1]{label=RQ\arabic*.,ref=RQ\arabic*}

\acrodef{sl}[SL]{Sign Language}
\acrodef{slt}[SLT]{Sign Language Translation}
\acrodef{sls}[SLS]{Sign Language Synthesis}
\acrodef{slr}[SLR]{Sign Language Recognition}
\acrodef{mt}[MT]{Machine Translation}
\acrodef{rbmt}[RBMT]{Rule-based Machine Translation}
\acrodef{smt}[SMT]{Statistical Machine Translation}
\acrodef{nmt}[NMT]{Neural Machine Translation}
\acrodef{vgt}[VGT]{Flemish Sign Language}
\acrodef{asl}[ASL]{American Sign Language}
\acrodef{bsl}[BSL]{British Sign Language}
\acrodef{dgs}[DGS]{German Sign Language}
\acrodef{csl}[CSL]{Chinese Sign Language}
\acrodef{ksl}[KSL]{Korean Sign Language}
\acrodef{lsfb}[LSFB]{French Belgian Sign Language}
\acrodef{ngt}[NGT]{Sign Language of the Netherlands}
\acrodef{slc}[SLC]{sign language community}
\acrodefplural{slc}[SLCs]{sign language communities}
\acrodef{cnn}[CNN]{Convolutional Neural Network}
\acrodefplural{cnn}[CNNs]{Convolutional Neural Networks}
\acrodef{rnn}[RNN]{Recurrent Neural Network}
\acrodefplural{rnn}[RNNs]{Recurrent Neural Networks}
\acrodef{cslr}[CSLR]{Continuous Sign Language Recognition}
\acrodef{stmc}[STMC]{Spatio-Temporal Multi-Cue}
\acrodef{lstm}[LSTM]{Long Short-Term Memory}
\acrodefplural{lstm}[LSTMs]{Long Short-Term Memory Networks}
\acrodef{tin}[TIN]{Temporal Inception Networks}
\acrodef{fps}[FPS]{Frames Per Second}
\acrodef{fsdc}[FSDC]{Frame Stream Density Compression}
\acrodef{dhbigru}[DH-BiGRU]{Dynamic Hierarchical Bidirectional GRU}
\acrodef{gdpr}[GDPR]{General Data Protection Regulation}
\acrodef{sota}[SOTA]{state of the art}
\acrodef{LSTM}[LSTM]{Long-Short Term Memory}
\acrodef{GRU}[GRU]{Gated Recurrent Unit}
\acrodef{rrg}[RRG]{role and reference grammar}
\begin{document}
\title{Machine Translation from Signed to Spoken Languages: State of the Art and Challenges}
\author{Mathieu De Coster \and Dimitar Shterionov \and Mieke Van Herreweghe \and Joni Dambre}
\date{}

\maketitle

\begin{abstract}
Automatic translation from signed to spoken languages is an interdisciplinary research domain on the intersection of computer vision, machine translation (MT), and linguistics.
While the domain is growing in terms of popularity --- the majority of scientific papers on sign language (SL) translation have been published in the past three years --- research in this domain is performed mostly by computer scientists in isolation. This article presents a critical overview of the work on SL translation. We first give a high level introduction to SL linguistics and MT to illustrate the requirements of automatic SL translation. Then, we present a systematic literature review of the state of the art in the domain. Finally, we outline important challenges for future research.
We find that significant advances have been made on the shoulders of spoken language MT research. However, current approaches often lack linguistic motivation or are not adapted to the different characteristics of SLs. We explore challenges related to the representation of SL data, the collection of datasets, the need for interdisciplinary research, and requirements for moving beyond research, towards applications.
We advocate for interdisciplinary research and for grounding future research in linguistic analysis of SLs.
Furthermore, the inclusion of deaf \emph{and} hearing end users of SL translation applications in use case identification, data collection, and evaluation, is of the utmost importance in the creation of useful SL translation models.
\end{abstract}

\section{Introduction}
\subsection{Scope of this article}
The speedy progress in deep learning has seemingly enabled a bevy of new applications
related to sign language recognition,
translation, and synthesis, which can be grouped under the umbrella term ``sign language processing.''
Sign language recognition
can be likened to ``information extraction from sign language data,'' for example
fingerspelling recognition and sign classification. \ac{slt} maps
this extracted information to meaning and translates it to another (signed or spoken) language;
the opposite direction, from text to sign language, is also possible.
\ac{sls} aims to generate sign language from some representation of meaning, for example through virtual
avatars or by stitching together prerecorded videos, each of which is associated with a specific sign or sign sequence. In this article, we are zooming in on
translation from signed languages to spoken languages.

In particular, we focus on translating videos containing sign language conversations to text, i.e., the written form of spoken language.
We will only discuss \ac{slt} models that support \emph{video data} as input, as opposed to models that require wearable bracelets or gloves, or 3D cameras. This choice is motivated by the fact that any system designed to be used
in an everyday setting cannot be expected to be intrusive. Systems
that use smart gloves, wristbands or other wearables \emph{are} intrusive and are unable to capture all information
present in signing, such as nonmanual actions. They are
not usable nor accepted by \acp{slc} \cite{erard2017sign}.
% Smart wearables are luckily not \colorbox{yellow}{a necessity}, either.
Humans can understand sign language through visual observation and many people have access to a camera at any time
through their smartphone. An \ac{slt} system
built to work with RGB videos therefore seems to be technologically and economically feasible, and potentially
user-friendly.

\begin{figure}
\centering
\includegraphics[width=0.45\textwidth]{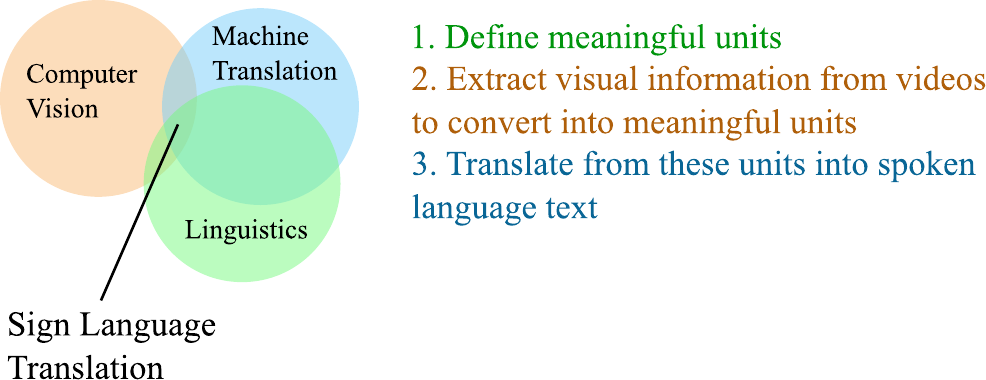}
\caption{Sign language translation lies on the intersection of computer vision, machine translation, and linguistics. Each of these domains tackles different related challenges and interdisciplinary research is required to solve sign language translation.}
\label{fig:slt_scheme}
\end{figure}

\subsection{An interdisciplinary challenge}
Before we delve into the domains of sign language recognition and translation, we need to address some misconceptions about what a sign language translation model is.
Several previously published scientific papers oversimplify this
domain, likening sign language recognition to gesture recognition,
or even presenting a fingerspelling system as an \ac{slt} solution. Such classifications
are overly simplified and incorrect. They may lead to a misunderstanding
of the technical challenges that must be solved.
Fig.~\ref{fig:slt_scheme} positions \ac{slt} on the intersection of computer vision, machine translation, and linguistics. Experts from each domain must come together to truly address sign language translation.

A crucial part of sign language research in a computational context, is understanding the difference between sign language \emph{recognition} on the one hand and sign language \emph{translation} on the other. The line between them is sometimes blurred in scientific papers. Sign language recognition is typically divided into \emph{isolated} and \emph{continuous} recognition. In isolated sign recognition, individual signs are classified, e.g., from a video input to a single sign. Continuous sign language recognition instead considers sequences of two or more signs. Sign language recognition can for example be applied to transcribe sign language videos to sign language glosses.
Both the input (videos) and the output (glosses) are related to the same language. In sign language translation, however, sequences of signs are \emph{translated} into a \emph{different} language.

\subsection{Research questions}
This article aims to provide a comprehensive overview of the
\ac{sota} of signed to spoken language translation. To do this, we perform a systematic
literature review and discuss the state of the domain. We aim to find answers to the following research questions.
\begin{questions}
    \item How should we represent sign language data for \ac{mt} purposes?\label{rq:sl_representation}
    \item Which algorithms are currently the \ac{sota} for \ac{slt}?\label{rq:slt_sota}
    \item Which datasets are used, for which languages, and what are the properties of these datasets?\label{rq:datasets}
    \item How are current \ac{slt} models evaluated and is this sufficient?\label{rq:evaluation}
\end{questions}
Furthermore, our goal is to
list several challenges in \ac{slt} that are often overlooked. These
challenges are of a technical and linguistic nature. We propose research directions to tackle these challenges.

Therefore, this article is not only
an introduction for researchers taking their first steps in the domain of \ac{slt},
but also a compass to guide future research.

\subsection{Structure of this article}
We discuss the inclusion criteria and search strategy for our systematic literature search in Section~\ref{sec:methodology}. Then, we provide a high level overview of some required background information on sign
languages and machine translation in Section~\ref{sec:signlanguages} and Section~\ref{sec:mt},
respectively. We objectively compare the results of the considered papers on \ac{slt} in Section~\ref{sec:literature};
this includes a section
focusing on a specific benchmark dataset in Section~\ref{sec:benchmark}.
The findings of the literature overview are summarized and discussed in Section~\ref{sec:discussion}.
We base ourselves on this discussion to present several outstanding challenges in \ac{slt} in Section~\ref{sec:challenges}.
The conclusion and takeaway messages are given in Section~\ref{sec:conclusion}.

\section{Literature review methodology}
\label{sec:methodology}
\subsection{Inclusion criteria and search strategy}
To provide an overview of sound \ac{slt} research, we adhere to the following principles in our literature search. We consider only peer reviewed publications. We include journal articles as well as conference papers: the latter are especially important in computer science research. Of course, any paper that is included must be on the topic of sign language machine translation and must not misrepresent the natural language status of sign languages. Therefore, we omit any papers that present classification of individual signs or fingerspelling recognition as sign language translation models. As we focus on nonintrusive translation from sign languages to text, we exclude papers that use gloves or other wearable devices. Finally, we emphasize the importance of (ethically) correct language use. We do not consider papers that make use of the terms ``deaf/dumb'' or ``deaf/mute''\footnote{Both of these terms can be found to be offensive, as explained by the United States National Association of the Deaf (\url{https://www.nad.org/resources/american-sign-language/community-and-culture-frequently-asked-questions/}).}, or that imply that sign language users are not ``normal'', or that they have problems communicating or need help living their daily lives.

These principles are summarized into the following inclusion criteria. Any paper that is included in our literature overview must:
\begin{itemize}
    \item be written in English,
    \item be peer reviewed,
    \item propose, implement and evaluate a sign language machine translation system from a sign language to a spoken language,
    \item present a nonintrusive system based {\em only} on RGB camera inputs,
    \item not use terms that can be construed as offensive to members of sign language communities.
\end{itemize}

Three scientific databases were queried: Google Scholar, Web of
Science and IEEE Xplore\footnote{Google Scholar: \url{https://scholar.google.com}, Web of Science: \url{https://www.webofscience.com}, IEEE Xplore: \url{https://ieeexplore.ieee.org/}.}. Four queries were used to obtain initial results: ``sign language translation'',
``sign language machine translation'', ``gloss translation'' and ``gloss machine translation''.
These key phrases were chosen for the following reasons.
We want to obtain scientific research papers on the topic of \ac{mt} from signed to spoken languages: therefore we search for ``sign language machine translation''. Several papers perform translation between sign language glosses and spoken language text (as we will discuss), hence ``gloss machine translation''. As many papers omit the word ``machine'' in ``machine translation'', we also include the key phrases ``sign language translation'' and ``gloss translation''.

The initial search, executed on August 19, 2021,
yielded {\bf 716 results}. 239 duplicate entries were removed, leaving 477 papers. In addition, 10
non-English papers were removed. 

Title screening was used next on the remaining {\bf 467 papers}. First we checked the intrusiveness criterion based on titles: any paper addressing \ac{slt} with gloves or other wearable sensors was removed. This was done by listing all paper titles containing the words
``glove'', ``armband'' or ``wearable'' and manually removing those papers proposing an intrusive system.
After this filtering step, {\bf 453 papers} remained. 

Subsequently, we screened the titles to exclude papers that do not address automatic translation in the direction of signed to spoken languages. We removed papers on human translation and interpretation, papers on translation from spoken to signed languages and papers not related to sign languages at all\footnote{The search query ``sign translation'' also yields papers on the recognition of traffic signs and other public signage.}. As a consequence, 256 papers were removed.

For the remaining {\bf 197 papers}, title and abstract were analyzed more thoroughly to identify remaining mismatches with our inclusion criteria. Papers were deemed irrelevant for several possible reasons. They could for example discuss the topics of fingerspelling recognition (39 papers),
isolated sign recognition (31 papers) or continuous sign language recognition (6 papers).
Other papers were removed because they considered intrusive methods but passed the earlier title based screening
(20 papers). 10 papers did not propose a machine translation system and were also removed. 36 papers
were removed because they only considered translation from spoken languages to signed languages (these papers had 
passed the title based screening).

16 papers used terms such as ``deaf/dumb'' or ``abnormal'' to refer to sign language users and therefore these papers are excluded from our overview.  5 more papers were removed for various other reasons.
Note that some papers consider both fingerspelling recognition and isolated sign recognition, hence
they are counted twice. 

After this title and abstract based screening, {\bf 55 papers} remained. These papers
were read in full to determine their relevance. We excluded papers that
do not describe a translation model or formulate a proposal only, i.e., papers that do not present any methodology or results. 

After all exclusion steps, the remaining {\bf 32 papers},  4.5\%  of the original 716 search results, are discussed in this work. These are peer reviewed papers in English that propose, implement and
evaluate a sign language machine translation system from a sign language to a spoken language using an RGB
camera and do not contain descriptions that are explicitly offensive to members of the sign language communities.

The list of all 716 search results, as well as the list of the 32 selected papers is provided as supplementary material (Resource 1).

\section{Sign language background}
\label{sec:signlanguages}
\subsection{Signed language}

It is a common misconception that there exists a single, universal, sign language. Just like spoken
languages, sign languages evolve naturally through time and space. Several countries
have national sign languages, but often there are also regional differences and local dialects.
Furthermore, signs in a sign language do not have a one-to-one mapping to words in any spoken language:
translation is not as simple as recognizing individual signs and replacing them with the corresponding
words in a spoken language. In summary: sign languages have distinct vocabularies and grammars and they are not tied to any spoken language. Even in two regions with a shared spoken language,
the regional sign languages used can differ greatly.
In the Netherlands and in Flanders (Belgium), for example, the majority spoken language is
Dutch. However, \ac{vgt} and the \ac{ngt} are quite different. Meanwhile, \ac{vgt}
is linguistically and historically much closer to \ac{lsfb} \cite{vermeerbergen201336},
the sign language used primarily in the French-speaking part of Belgium, because both
originate from a common Belgian Sign Language, diverging in the 1990s \cite{van2009flemish}. In a similar vein, \ac{asl} and \ac{bsl} are completely different even though the two countries share a spoken language, i.e., English.

\subsection{A high level overview of components of sign languages}
We now provide a very high level overview of sign languages from a linguistic point of view.
It is by no means comprehensive, but it illustrates why \ac{slt} is much broader than gesture recognition and even sign recognition. Remember that there is no single universal sign
language, and therefore some of the notions that we discuss here may not apply to all sign languages.

Sign languages are visual; they make use of a large space around the signer. Signs are not
composed solely out of manual gestures. In fact, there are many more
components to a sign. Stokoe stated in 1960 that signs are composed of hand shape, movement and
place of articulation parameters \cite{stokoe1960sign}. Battison later added orientation, both of the palm and of the fingers \cite{battison1978lexical}.
There are also nonmanual components such as mouth patterns. Mouth patterns can be divided into mouthings --- where the pattern refers to (part of) a spoken language word --- and mouth gestures, e.g., touting one's lips. Nonmanual components play an important role in sign language lexicons and grammars \cite{bank2011variation}. They can for example
separate minimal pairs: these are signs which share all articulation parameters apart from one.
When hand shape, orientation, movement and place of articulation are identical, mouth patterns can for example be used to differentiate two signs. Nonmanual actions are not only important at the lexical level as just illustrated, but also at the grammatical level. A clear example of this can be found in eyebrow movements: furrowing or raising
the eyebrows can signal that a question is being asked, as well as indicate the type of question (open or closed).

Sign languages exhibit simultaneity on several levels.
There is simultaneity on the component level:
as explained above, manual actions can be combined with nonmanual actions simultaneously. We also
observe simultaneity at the utterance level. It is, for example, possible to turn a positive
utterance into a negative utterance by shaking one's head while performing the manual actions.
Another example is the use of eyebrow movements to transform a statement into a question.

The space around the signer can also be utilized to indicate for instance the location or moment in time
of the conversational topic. A signer can point behind their back to specify that an event occurred
in the past and likewise, point in front of them to indicate a future event. An imaginary
timeline can also be constructed in front of the signer, with time passing from left to right.
Space is also used to position referents \cite{vermeerbergen201336,perniss201219}.
For example, a person can be discussing a conversation
with their mother and father. Both referents get assigned a location (locus) in the signing space
and further references to these persons are made by pointing to, looking at, or signing towards
these loci. For example, ``mom gives something to dad'' can be signed by moving the sign for ``to give''
from the locus associated with the mother to the one associated with the father.
Modeling space, detecting positions in space,
and remembering these positions is important for \ac{slt} models.

Another important aspect of sign languages is the use of classifiers. Zwitserlood describes them as
``morphemes with a nonspecific meaning, which are expressed by particular configurations of the manual
articulator (or: hands) and which represent entities by denoting salient characteristics''
\cite{zwitserlood2012classifiers}. There are many more intricacies of classifiers than can be
listed here, so we give a limited set of examples instead.
Several types of classifiers exist. They can for example
represent nouns or adjectives according to their shape or size.
Whole entity classifiers can be used to represent objects, e.g., a flat hand can represent a car;
handling classifiers can be used to indicate that an object is being handled, e.g., a pencil
is picked up from a table. In a whole entity classifier, the articulator \emph{is} the object, whereas
in a handling classifier it \emph{operates on} the object.

The vocabularies of sign languages are not fixed. Oftentimes new signs are constructed
by sign language users. On the one hand, sign languages can borrow signs from other
sign languages, similar to loanwords in spoken languages. In this case, these signs
are part of the \emph{established} lexicon. On the other hand, there is the \emph{productive}
lexicon --- one can create an ad hoc sign.
Vermeerbergen
gives the example of ``a man walking on long legs''
in \ac{vgt}: rather than expressing
this clause by signing ``man'', ``walk'', ``long'' and ``legs'', the hands are used
(as classifiers) to imitate the man walking \cite{vermeerbergen2006past}.
Both the established and productive lexicons are integral parts of sign languages.

Fingerspelling can be used to convey concepts for which a sign does not (yet) exist, or to introduce a person
who has not yet been assigned a name sign. It is based on the alphabet of a spoken language, where every
letter in that alphabet has a corresponding (static or dynamic) sign. Fingerspelling is also not shared between sign languages. For example, in \ac{asl}, fingerspelling is one handed, but in \ac{bsl} two hands are used.

We have now discussed seven important aspects of signing: manual actions, nonmanual actions,
signing space, classifiers, the productive lexicon, simultaneity and fingerspelling. Models for \ac{slt} require
the ability to deal with all of these aspects in some way, either explicitly or implicitly.

These aspects cannot be trivially extracted from sign language videos as they are. The videos first need to be processed into some representation of sign language. This representation can be written, graphical or computational. No matter which kind is used, it needs to contain information on the aforementioned aspects to allow for translation to different languages.

\subsection{Notation systems for sign languages}
\label{sec:notation}
Unlike many spoken languages, sign languages do not have a standardized written form. Several notation systems do exist, but none of them are generally accepted as a standard~\cite{FrishbergHoitingSlobin_SignTranscription}.
The earliest notation system was proposed in the 1960s by Stokoe: the Stokoe notation~\cite{stokoe1960sign}. It was designed for \ac{asl} and comprises a set of symbols to notate the different components of signs.
The
position, movement and orientation of the hands are encoded in iconic symbols, and for hand shapes,
letters from the Latin alphabet corresponding to the most similar fingerspelling hand shape are used \cite{stokoe1960sign}. Later, in the 1970s,
Sutton introduced SignWriting\footnote{\url{https://signwriting.org/}}: a notation system for sign languages based on a dance choreography notation system \cite{sutton1981sign}. The SignWriting notation for a sign is composed of iconic symbols for the hands, face and body. The signing location and movements are also encoded in symbols, in order to capture the dynamic nature of signing.
SignWriting is designed as a system for writing signed utterances
for everyday communication.
In 1989, the Hamburg Notation System (HamNoSys) was introduced \cite{Prillwitz1989hamnosys}. Unlike SignWriting, it is designed mainly
for linguistic analysis of sign languages.
It encodes hand shapes, hand orientation, movements and nonmanual components in the form of symbols.

Stokoe notation, SignWriting and HamNoSys represent the visual nature of signs in a compact format. They are notation systems that operate on the phonological level. These systems, however, do not capture the meaning of signs. In linguistic analysis of sign languages, glosses are typically used to represent meaning.
A sign language gloss is a written representation of a sign in one or more words of a spoken language, commonly the majority language of the region.
Glosses can be composed of single words in the spoken language, but also of combinations of words.
Examples of glosses are: ``CAR'', ``BRIDGE'', but also ``car-crosses-bridge''.
Glosses do not accurately represent the meaning of signs in all cases and glossing
has several limitations and problems \cite{FrishbergHoitingSlobin_SignTranscription}.
They are inherently sequential, whereas signs often exhibit
simultaneity \cite{vermeerbergen2007simultaneity}\footnote{For this reason, annotators of sign language corpora often provide two parallel gloss tiers: one per hand.}. Furthermore, as glosses are based
on spoken languages, there may be an implicit influence of the spoken language projected onto the sign
language \cite{vermeerbergen2006past,FrishbergHoitingSlobin_SignTranscription}. Finally, there is no universal standard on how glosses should be
constructed: this leads to differences between corpora of different sign languages, or even between
several sign language annotators working on the same corpus.

Sign\_A is a recently developed framework aiming to define an architecture that is sufficiently robust to model sign languages on
both the phonological level as well as containing meaning (when combined with a \ac{rrg}) \cite{Sign_A_Thesis}. Sign\_A with \ac{rrg} does not only encode the meaning of sign language utterances, but also parameters
pertaining to manual and nonmanual actions. De Sisto \emph{et al.} propose investigating the
application of Sign\_A for data-driven \ac{slt} systems~\cite{de-sisto-etal-2021-defining}.

The above notation systems for sign languages range from graphical to written and computational representations of signs and signed utterances. None of these notation systems were originally designed for the purpose of automatic translation from signed to spoken languages and, in fact, only glosses are currently used for \ac{slt}.
One reason is that sign glosses are similar on several levels to spoken language words, facilitating translation using spoken language \ac{mt} techniques.

\section{Machine translation}
\label{sec:mt}
\subsection{Spoken language MT}
Machine translation is a sequence to sequence task. That is, given an input sequence of tokens that constitute a sentence in a source language, an \ac{mt} system generates a new sequence of tokens that represent a sentence in a target language. In fact, as \ac{mt} is a probabilistic task, the generated sequence is the most likely translation of the input sequence\footnote{Corpus-based paradigms, such as statistical and neural \ac{mt}, rely on statistics derived from the alignment of source and target sentences in the training corpora. Rule-based \ac{mt}, one of the earliest paradigms, is based on linguistically motivated rules and dictionaries.}. A token refers to a sentence construction unit: a word, a number, a symbol, a character or a subword unit. 

Current \ac{sota} models for spoken language \ac{mt} are based on a neural encoder-decoder architecture: (i) an encoder network encodes an input sequence in the source language into a multi-dimensional representation; (ii) it is then fed into a decoder network which generates a hypothesis translation conditioned on this representation. The original encoder-decoder was based on \acp{rnn}~\cite{sutskever2014sequence}. To deal with long sequences, \acp{lstm}~\cite{Hochreiter1997LSTM} and \acp{GRU}~\cite{cho2014neural} were used. To further improve the performance of RNN-based MT, an attention mechanism was introduced by Bahdanau \emph{et al.} \cite{bahdanau2015neural}. In recent years the transformer architecture~\cite{vaswani2017attention}, based primarily on the idea of attention (in combination with positional encoding) has pushed the \ac{sota} even further.

As noted above, a sentence is broken down into tokens and each token is fed into the \ac{nmt} model. \ac{rnn}-based models process a sequence one token at a time, whereas transformer based models operate on multiple tokens in parallel. Regardless of the architecture type, \ac{nmt} converts each token into a multidimensional representation before that token representation is used in the encoder or decoder to construct a sentence level representation. These token representations, typically referred to as \emph{word embeddings}, encode the meaning of a token based on its context and can be learned along with the training of an \ac{nmt} model, or independently and used to bootstrap the \ac{nmt} training. Learning word embeddings is a monolingual task, since they are associated with tokens in a particular language. Given that for a large number of languages and use cases monolingual data is abundant, it is relatively easy to build word embedding models of high quality and coverage. Building such word embedding models is typically performed using unsupervised algorithms such as GLoVe \cite{pennington2014glove}, BERT \cite{devlin2018bert} and BART~\cite{Lewis2020Bart}. These algorithms encode words into vectors in such a way that the vectors of related words are similar\footnote{According to the Distributional Semantics, words that have the same or similar meaning appear in the same context and as such the meaning of a word can be defined by the context in which it appears~\cite{Harris1954_DistributionalStructure,Firth1957}.}.

The domain of spoken language \ac{mt} is extensive and the current \ac{sota} of \ac{nmt} builds upon years of research. To provide a complete overview of spoken language \ac{mt} is out of scope for our article.
For a
more in depth overview of the domain, we refer readers to the work of Stahlberg \cite{stahlberg2020neural}.

\begin{figure}
\centering
\includegraphics[width=0.95\textwidth]{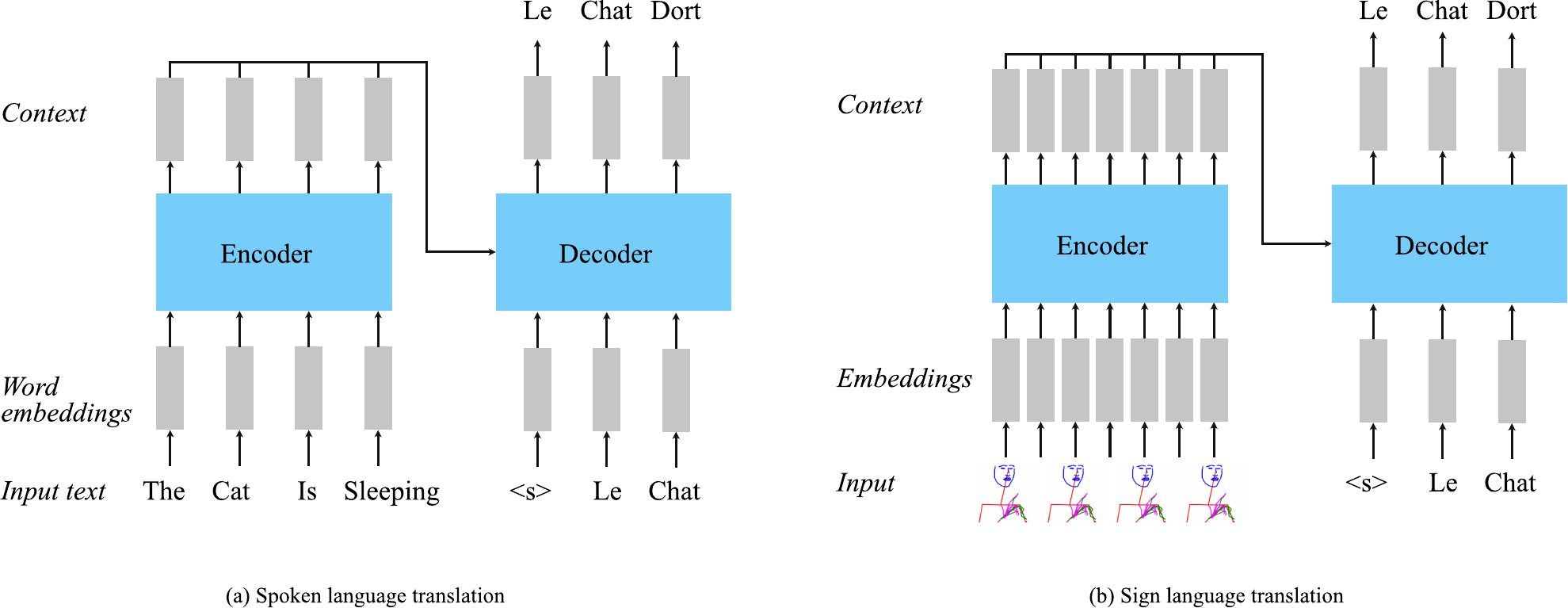}
\caption{Neural machine translation models for spoken (a) and sign (b) language translation are similar; the
main difference is the input modality. In this case, the sign language representation is illustrated as human pose estimation keypoints. The pose illustration is adapted from~\cite{de2020sign}.}
\label{fig:machinetrans}
\end{figure}

\subsection{Sign language MT}\label{slt}
Conceptually, sign language \ac{mt} and spoken language \ac{mt} are similar. The main difference is the input modality.
Spoken language \ac{mt} operates on two streams of discrete tokens (text to text). As sign languages
have no standardized notation system, a generic \ac{slt} model must translate from a continuous stream to a discrete stream (video to text). To reduce the complexity of this problem, sign language videos are discretized to a sequence of still frames that make up the video. \ac{slt} can now be framed as a sequence-to-sequence, frame-to-token task. As they are, these individual frames do not convey meaning in the way that the word embeddings in a spoken language translation model do. Notwithstanding that it is possible to train \ac{slt} models using frame based representations as inputs, the extraction of salient sign language representations is required to facilitate the modeling of meaning in sign language encoders.

Following the encoder-decoder \ac{nmt} architecture, the first step in \ac{slt} would be the encoding of sign language sentences captured in videos, broken down into frames. Encoding a sign language sentence is a challenge on its own. Consider, for example, that each video captures not only the signer but also the background, other signers and other noise, in general. Focusing on the sign language information from the video is essential for the translation task.
Therefore, before encoding, such a sentence should be processed and the sign language information isolated. This is the extraction of the sign language representation mentioned above. This process is called ``sign language recognition'' and it is performed before translation.
Whereas spoken language texts can be divided into words, subwords, characters or other discrete tokens, which
can be mapped to real-valued vectors, i.e., word embeddings,
this is more difficult to do for
sign language videos. There is no explicit segmentation between individual signs, for example.
In deep learning, typically, \acp{cnn} are used to extract information from videos.
These videos can be processed either frame by frame or multiple frames at a time. In the former case, only spatial information is captured, with temporal information implicitly encoded in the sequence.

In the latter technique, both spatial and temporal information is captured.

Fig.~\ref{fig:machinetrans} shows a spoken language \ac{nmt} and sign language \ac{nmt} model side by side. The main difference between the two is the input modality. For a spoken language \ac{nmt}
model, both the inputs and outputs are text. For a sign language \ac{nmt} model, the inputs are some
representation of sign language (in the case of this illustration, per-frame human pose keypoints extracted with
OpenPose \cite{cao2019openpose}).

\subsubsection{Sign language representations}
For the encoder of the translation
model to capture the meaning of the sign language utterance, a salient representation for sign language videos is required. We can differentiate between: (i) representations that are linked to the source modality, namely videos, and (ii) linguistically motivated representations.

As will be discussed in Section~\ref{sec:lit_repr}, the former type of representations are often frame based, i.e., every frame in the video is assigned a vector, or clip based, i.e., clips of arbitrary length are assigned a vector. These type of representations are rather simple to derive, e.g., by extracting information directly from the \ac{cnn}. However, they suffer from two main drawbacks. First, such representations are fairly long. For example, the RWTH-PHOENIX-Weather 2014T dataset \cite{camgoz2018neural}
contains samples of on average 114 frames (in \ac{dgs}),
whereas the average sentence length (in German) is 13.7 words in that dataset.
As a result, frame based representations for sign languages negatively impact the computational
performance of \ac{slt} models. Second, such representations do not originate from domain knowledge. That is, they capture neither the syntax nor semantics of sign language. If semantic information is not encoded in the sign language representation, the translation model is forced to model the semantics and perform translation at the same time. If, in contrast, semantic information is encoded in
the representation, only translation needs to be learned, which is an easier task.

The second category includes a range of linguistically motivated representations, from semantic representations to individual sign representations.
In Section~\ref{sec:notation} we presented an overview of some notation systems for sign languages: Stokoe notation, SignWriting, HamNoSys, glosses, and Sign\_A. These notation systems can be used as
representations in an \ac{slt} model. From the aforementioned notation systems,
only glosses are used in current \ac{sota} models.
Therefore in the remainder of this article, our discussion on the application of preexisting notation systems in \ac{slt} will be limited to glosses.

\subsubsection{Tasks}
\label{sec:slt_tasks}

\begin{figure}
\centering
\includegraphics[width=0.95\textwidth]{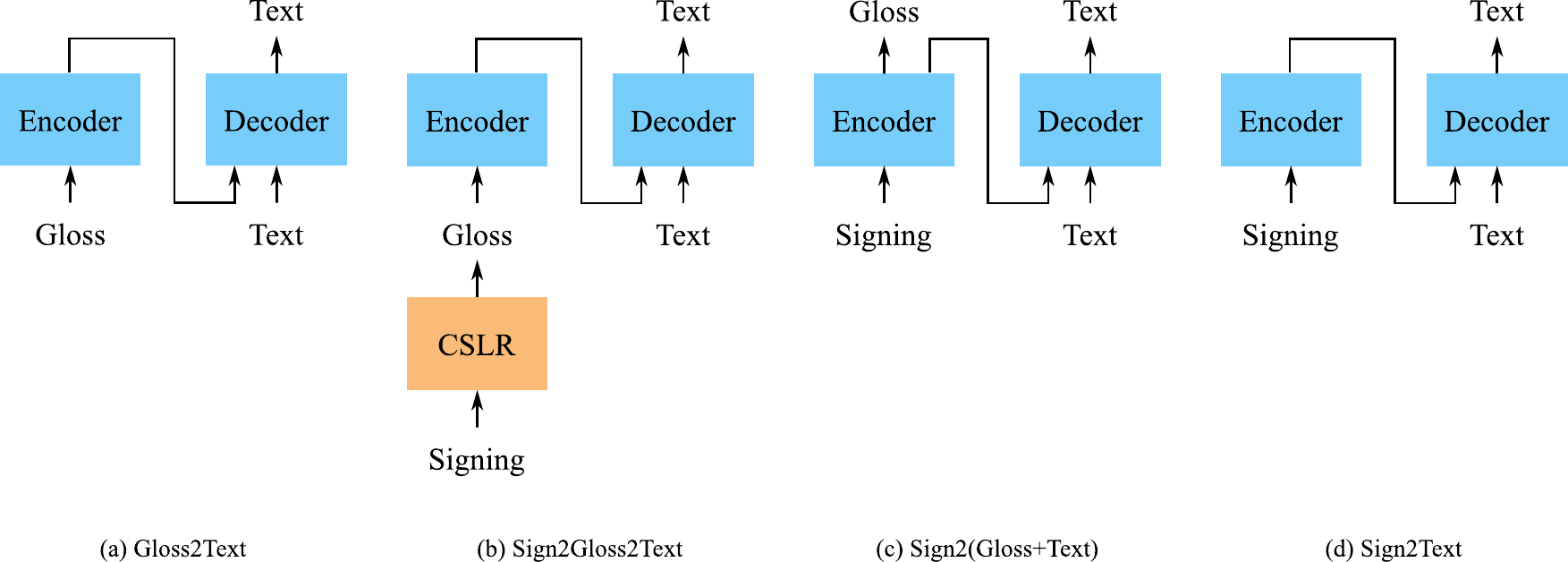}
\caption{We find four distinct translation tasks in the considered scientific literature on \ac{slt}. Each task has different inputs and/or outputs and implications on the required labels and resulting scores on translation metrics. CSLR: Continuous sign language recognition.}
\label{fig:task_diagrams}
\end{figure}

The reviewed papers cover four distinct translation tasks
that can be classified based on whether, and how, glosses are used.
To denote these tasks, we borrow the naming conventions from Camgoz \emph{et al.} \cite{camgoz2020sign}.
First, translation from sign language
glosses to spoken language text is considered (Gloss2Text); such
a system assumes the existence of a perfect sign language recognizer to
perform the transformation from sign language videos into sign language glosses.
The three other tasks consider video inputs. Sign to text translation
translates directly from sign language videos to text (Sign2Text). Sign to gloss to text translation first
converts the sign language video into glosses through sign language recognition methods and then translates
these glosses into text (Sign2Gloss2Text). Finally,
the task to jointly predict glosses and translate into spoken language text is called Sign2(Gloss+Text).
We provide diagrams of these tasks in Fig.~\ref{fig:task_diagrams}.

\paragraph{Gloss2Text} Gloss2Text models provide a reference for the performance that can be achieved using a salient representation. Therefore they can serve as a compass for the design of sign language representations and the corresponding sign language recognition systems. Note once more
that glosses do not capture all linguistic properties of signs; therefore the performance of a Gloss2Text model is not an upper bound on the performance of an \ac{slt} model.

\paragraph{Sign2Gloss2Text} A Sign2Gloss2Text translation system includes the sign language recognition
system as the first step. Consequently,
errors made by the recognition system are propagated to the translation
system. Camgoz \emph{et al.} for example report a drop in translation accuracy when comparing
a Sign2Gloss2Text system to a Gloss2Text system \cite{camgoz2018neural}. However, such a drop in performance can be avoided by using salient sign language representations \cite{yin2020better}.
Note that a Sign2Gloss2Text translation system contains an information bottleneck
in the form of the gloss prediction. Furthermore, Sign2Gloss2Text models require a recognizer from signs to glosses at inference time.

\paragraph{Sign2(Gloss+Text)} Glosses can provide a supervised signal to a translation system
without being an information bottleneck, if the model is trained to jointly predict both glosses and text \cite{camgoz2020sign}. Such a model must be able to predict glosses and text from a single
sign language representation. The gloss labels provide additional information to the encoder,
facilitating the training process. 
In a Sign2Gloss2Text model, the translation models receives glosses as inputs: any information that is not present in glosses cannot be used to translate into spoken language text. In Sign2(Gloss+Text) models, however, the translation model input is the sign language representation.
In a Sign2Gloss2Text system, the sign language representation is exactly as salient as glosses. In a Sign2(Gloss+Text) system, the representation can contain additional information not present in glosses.
Another benefit of Sign2(Gloss+Text) models is that, after training (i.e., during inference), no gloss information is required: the model can be directly
applied to translate from the sign language representation into the spoken language text.

\paragraph{Sign2Text} Finally, Sign2Text performs both recognition and translation in an end-to-end set-up. It avoids the information
bottleneck presented by glosses as well as the need for gloss level annotations,
but requires a powerful sign language recognition system.
The creation of such recognition systems is currently heavily constrained by the limited amount of labeled data that is available.
Furthermore, research into sign language representations and sign language recognition is still ongoing.
Therefore, Sign2Text systems are often
outperformed by Gloss2Text, Sign2Gloss2Text and Sign2(Gloss+Text) systems:
this is discussed in Section~\ref{sec:benchmark}.

\subsection{Requirements for sign language MT}
\label{sec:reqs}
With the given information on sign language linguistics and \ac{mt} techniques, we are now able to sketch the
requirements for sign language \ac{mt}.

\subsubsection{Video processing and sign language representation} We need to be able to process sign language videos and convert them into an internal representation (sign language recognition). This representation must be rich enough to cover several aspects of sign languages (including manual and nonmanual actions, signing space, classifiers, the productive lexicon, simultaneity and fingerspelling). Ideally, this representation would be sign language agnostic, such that the sign language recognizer can be reused across languages. However, this is not a requirement for models designed for individual language pairs. We look in our literature overview for an answer to \ref{rq:sl_representation} on how we should represent sign language data.

\subsubsection{Translating between sign and spoken representations} We need to be able to translate from such a representation into a spoken language representation, which can be reused from existing spoken language \ac{mt} systems. We need to adapt \ac{nmt} systems to be able to work with the sign language representation, which will possibly contain simultaneous elements. By comparing different methods for \ac{slt}, we evaluate whether current algorithms are sufficient and which of them perform best in the current
\ac{sota} (\ref{rq:slt_sota}).

\subsubsection{Data requirements} To perform these operations using current \ac{sota} machine learning methods, we need large datasets. The collection of such datasets is expensive and should therefore be tailored to the wanted use cases. To determine these use cases, members of \acp{slc} must be involved. We answer \ref{rq:datasets} by providing an overview of existing datasets for \ac{slt}.

\section{Literature overview}
\label{sec:literature}
\subsection{Sign language MT}
Following our methodology on paper selection, laid out in Section~\ref{sec:methodology}, we obtain 32 papers published in the period from 2004 to August 2021. In the analysis we conduct, papers are classified based on tasks, datasets, methods and evaluation techniques.

\begin{figure}
\centering
\includegraphics[width=0.45\textwidth]{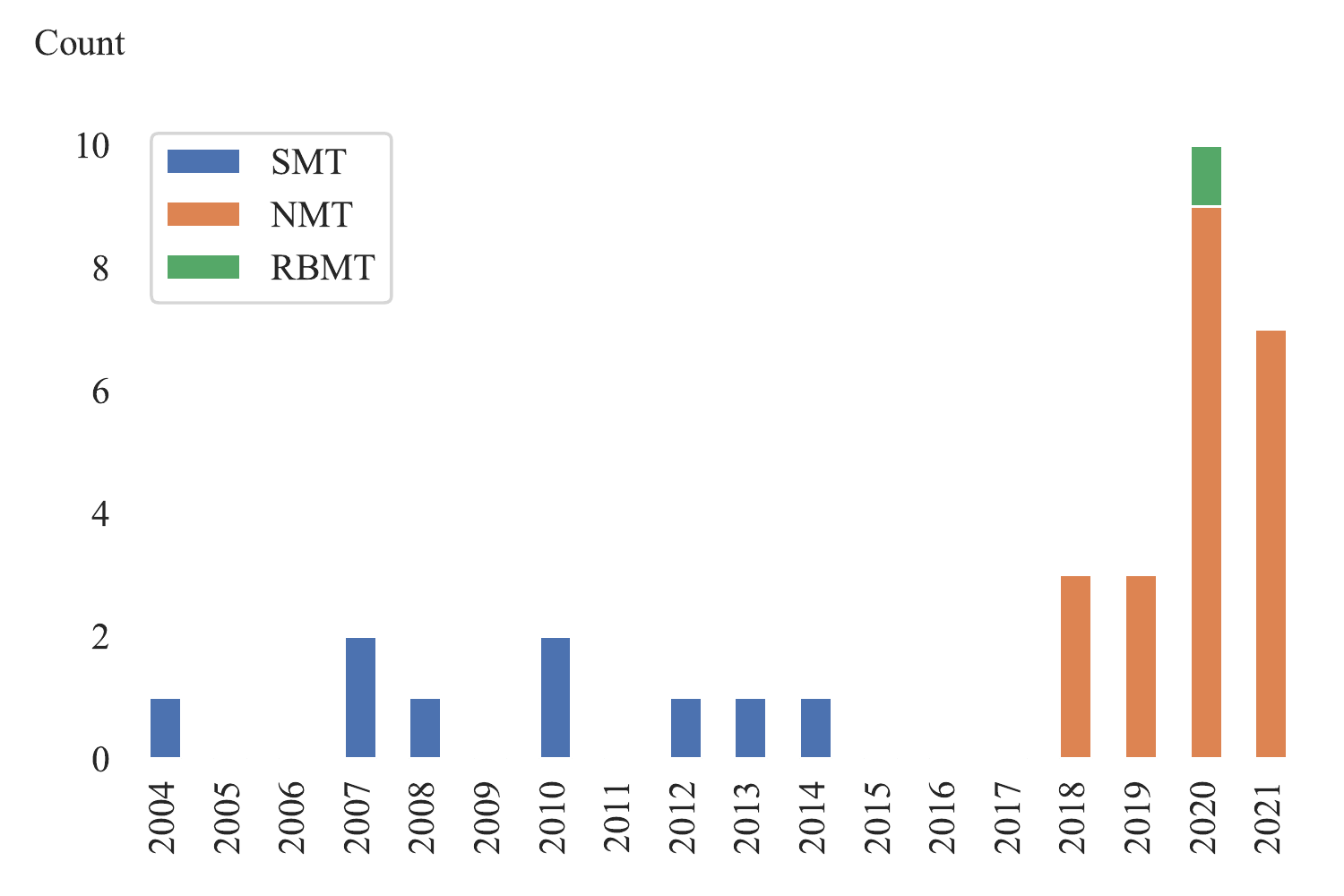}
\caption{The earlier papers on \ac{slt} all propose \ac{smt} models; since 2018, \ac{nmt} has become
the dominant variant. The considered papers were published between 2004 and 2021. SMT: Statistical Machine Translation, NMT: Neural Machine Translation, RBMT: Rule-based Machine Translation.}
\label{fig:years}
\end{figure}

Until the rise of deep neural models for \ac{mt} in 2017, most of the work on \ac{mt} from signed to spoken languages was based entirely on statistical methods~\cite{stein2012analysis,morrissey2007combining,forster2014extensions,stein2007hand,stein2010sign,lopez2010spanish,dreuw2008spoken,bungeroth2004statistical,schmidt2013using}. In 2018, the domain moved away from \ac{smt} and towards \ac{nmt}. This trend is
clearly visible in Fig.~\ref{fig:years}. This drastic shift was not only motivated by the successful applications
of \ac{nmt} techniques in spoken language \ac{mt}, but also by the publication of the RWTH-PHOENIX-Weather 2014T dataset
and the promising results obtained on that dataset using \ac{nmt} methods \cite{camgoz2018neural}. A single outlier is found in the paper by Luqman \emph{et al.}, who use \ac{rbmt} in 2020 to translate from Arabic sign language into Arabic~\cite{luqman2020machine}.

Between 2004 and 2018, research into translation from signed to spoken languages was sporadic (9 papers were published over 14 years). Since 2018, with the move towards \ac{nmt}, the domain has become more popular, with 23 papers in our subset published over the span of 3 years.

\subsection{Sign language representations}
\label{sec:lit_repr}
The sign language representations used in the current scientific literature range from glosses to representations extracted from videos.
In 2007 and 2008, before the widespread use of deep \acp{cnn} and while the majority of research was still being performed on Gloss2Text models,
two papers tackled Sign2Gloss2Text translation \cite{stein2007hand,dreuw2008spoken}. They both used
appearance based features in the form of 32 by 32 pixel grayscale images as well as motion based features
by performing hand tracking: the hand position, velocity and trajectory were used.
Schmidt \emph{et al.} perform Gloss2Text translation with an additional channel of visual information~\cite{schmidt2013using}. They track the position of the face and mouth using an active appearance model. Then, features are extracted from a landmark model of the face, such as eyebrow movements and mouth movements.
All other models from this period tackle Gloss2Text and consequently do not perform sign language feature extraction. All models since 2018 that include feature extraction, use neural networks to do so.

In modern \ac{slt}, we distinguish three feature extraction methods: 2D \acp{cnn}, human pose estimation (typically also performed with 2D \acp{cnn}), and 3D \acp{cnn}.
The most popular feature extraction method in modern \ac{slt} is the 2D \ac{cnn}.
Ten (53\% of 19) papers use a 2D \ac{cnn} as feature extractor \cite{camgoz2018neural,zheng2020improved,yin2020better,camgoz2020multi,orbay2020neural,camgoz2020sign,zhao2021conditional,zhou2021improving,zhou2021spatial,de-coster-etal-2021-frozen}.
Three of these use an additional 1D \ac{cnn} to temporally
process the resulting spatial features \cite{zheng2020improved,zhou2021improving,zhou2021spatial}.

Human pose estimation
systems are used to extract features in seven papers (37\%) \cite{ko2019neural,guo2019hierarchical,camgoz2020multi,orbay2020neural,kim2020robust,zhou2021spatial}. The estimated poses can be the sole inputs to the translation model \cite{ko2019neural,kim2020robust,orbay2020neural}, or augment
other spatial or spatio-temporal features \cite{guo2019hierarchical,camgoz2020multi,zhou2021spatial}.

The least popular deep feature extraction method is the 3D \ac{cnn}, with only three papers (16\%) using these
networks \cite{guo2018hierarchical,guo2019hierarchical,orbay2020neural}. Coincidentally, they
were also found to produce the least salient representations in a benchmarking study by Orbay \emph{et al.} \cite{orbay2020neural}, who compare different
feature extraction methods on the RWTH-PHOENIX-Weather 2014T dataset. They show that the best Sign2Text translation
performance (in terms of BLEU-4 score)
is achieved using 2D \acp{cnn} pretrained on hand shape classification, followed by pose estimation,
2D \acp{cnn} and finally 3D \acp{cnn}.

The representations discussed here are extracted from sign language videos. Many papers focus on manual actions or simply
consider the whole video frames as inputs.
Of the 19 papers that perform Sign2Gloss2Text, Sign2(Gloss+Text) or Sign2Text translation, four
(21\%) explicitly include nonmanual features such as mouth patterns or features extracted after cropping the face from
the image \cite{kumar2018time,yin2020better,camgoz2020multi,zhou2021spatial}. The other papers focus on hand appearance
features or use full frame images.

\subsection{Sign language translation models}
The current \ac{sota} in \ac{slt} is entirely based on encoder-decoder \ac{nmt} models.
\acp{rnn} are evaluated in 12 papers
\cite{yin2020better,zheng2020improved,guo2018hierarchical,guo2019hierarchical,rodriguez2021important,camgoz2018neural,ko2019neural,orbay2020neural,kumar2018time,arvanitis2019translation,rodriguez2020understanding,partaourides2020variational}
and transformers also in 12 papers
\cite{yin2020better,zhao2021conditional,moryossef-etal-2021-data,zhou2021improving,camgoz2020multi,camgoz2020sign,ko2019neural,orbay2020neural,kim2020robust,moe2020unsupervised,de-coster-etal-2021-frozen,zhang-duh-2021-approaching}.
Within the \ac{rnn} based models, several attention schemes are used: no attention, Luong attention
\cite{luong-etal-2015-effective} and Bahdanau attention \cite{bahdanau2015neural}.

To the best of our knowledge, there has been no systematic comparison
of \acp{rnn} and transformers across multiple tasks and datasets for \ac{slt}.
Within the reviewed papers, we nonetheless find some comparisons between architectures with varying results. The first comparison
is performed by Ko \emph{et al.}~\cite{ko2019neural}
on the KETI dataset (\ac{ksl}).
Whereas \acp{rnn} with Luong attention obtain the highest ROUGE score, transformer based networks perform better in terms of METEOR, BLEU and CIDEr scores.
\acp{rnn} without attention or with Bahdanau attention
are outperformed by the other variants on all reported metrics.
Another comparison is performed by Orbay \emph{et al.} \cite{orbay2020neural} on the
RWTH-PHOENIX-Weather 2014T dataset. They obtain different results: an \ac{rnn} with Bahdanau
attention outperforms both an \ac{rnn} with Luong attention and a transformer in terms of ROUGE and BLEU scores.
Yin \emph{et al.} compare an \ac{rnn} with Bahdanau attention, an \ac{rnn} with Luong attention and a transformer, also on the RWTH-PHOENIX-Weather 2014T dataset. They find that the transformer outperforms the \acp{rnn}, and that an \ac{rnn} with Luong attention outperforms one with Bahdanau attention \cite{yin2020better}.
Finally, Camgoz \emph{et al.} also compare \acp{rnn} and transformers~\cite{camgoz2020sign}.
They report a large increase in BLEU scores when using transformers,
compared to a previous paper using \acp{rnn} \cite{camgoz2018neural}. However, the comparison is between
models with different feature extractors: the impact of the architecture versus that of the feature extractors is unclear. It is likely that replacing a 2D \ac{cnn} pretrained on
ImageNet \cite{deng2009imagenet} image classification with one pretrained on \ac{cslr} will
result in a significant increase in performance, especially when the \ac{cslr} model was trained on data from
the same source (i.e., RWTH-PHOENIX-Weather 2014), as is the case here.

We can conclude that the choice of architecture depends on the dataset, sign language representation and translation task. We analyze the performance differences between transformers and \acp{rnn} on the RWTH-PHOENIX-Weather 2014T dataset in Section~\ref{sec:benchmark}.

\subsection{Tasks}
In total, 17 papers (53\%) report on a Gloss2Text model \cite{bungeroth2004statistical,morrissey2007combining,stein2010sign,lopez2010spanish,stein2012analysis,schmidt2013using,forster2014extensions,camgoz2018neural,ko2019neural,arvanitis2019translation,luqman2020machine,yin2020better,camgoz2020sign,moe2020unsupervised,partaourides2020variational,moryossef-etal-2021-data,zhang-duh-2021-approaching}. Sign2Gloss2Text
models are proposed in seven papers (22\%) \cite{stein2007hand,dreuw2008spoken,camgoz2018neural,kumar2018time,yin2020better,camgoz2020sign,zhou2021improving}. Sign2(Gloss+Text) models are found three times (11\%) within the reviewed papers \cite{camgoz2020sign,zhou2021spatial,de-coster-etal-2021-frozen} and
Sign2Text models 12 times (38\%) \cite{guo2018hierarchical,camgoz2018neural,ko2019neural,zheng2020improved,guo2019hierarchical,camgoz2020multi,orbay2020neural,kim2020robust,camgoz2020sign,rodriguez2020understanding,zhao2021conditional,rodriguez2021important,zhou2021improving}.

\begin{figure}
\centering
\includegraphics[width=0.45\textwidth]{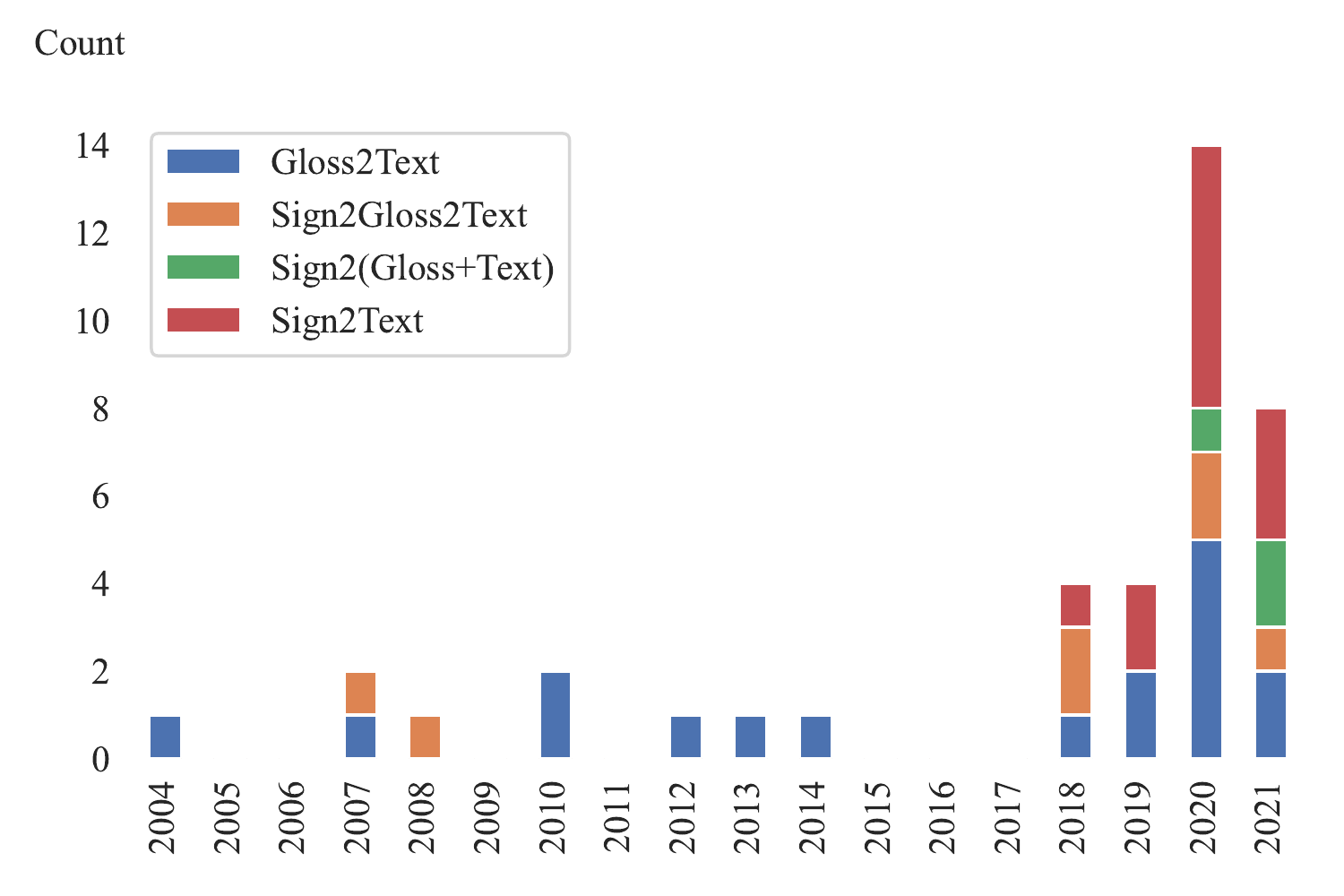}
\caption{Gloss-based models are used throughout the entire considered time period (2004-2021),
but since 2018 models which translate from video to text are gaining traction. As one paper may
discuss several tasks, the total count is higher than the amount of papers.}
\label{fig:tasks_per_year}
\end{figure}

We can distinguish two distinct eras in the \ac{slt} research domain: the era of \ac{smt} systems until 2015,
and the \ac{nmt} era starting in 2018. In the \ac{smt} era, Gloss2Text models were the most popular ones,
being proposed 7 times out of 9 models (78\%), the other two (22\%) being Sign2Gloss2Text models. In the
\ac{nmt} era, there is a shift in the distribution. Due to the availability of larger datasets,
deep neural feature extractors
and neural \ac{slr} models, Sign2Gloss2Text (17\% of 30), Sign2(Gloss+Text) (10\%) and Sign2Text (40\%) gain in popularity.
This gradual evolution from gloss based models towards end-to-end models is visible in Fig.~\ref{fig:tasks_per_year}.
At the same time, it is clear that the domain is still reliant on glosses. 60\% of the 30 models proposed since 2018
use gloss information in some form and 33\% of the proposed models since 2018 are Gloss2Text models.

\subsection{Datasets}
\begin{figure}
\centering
\includegraphics[width=0.45\textwidth]{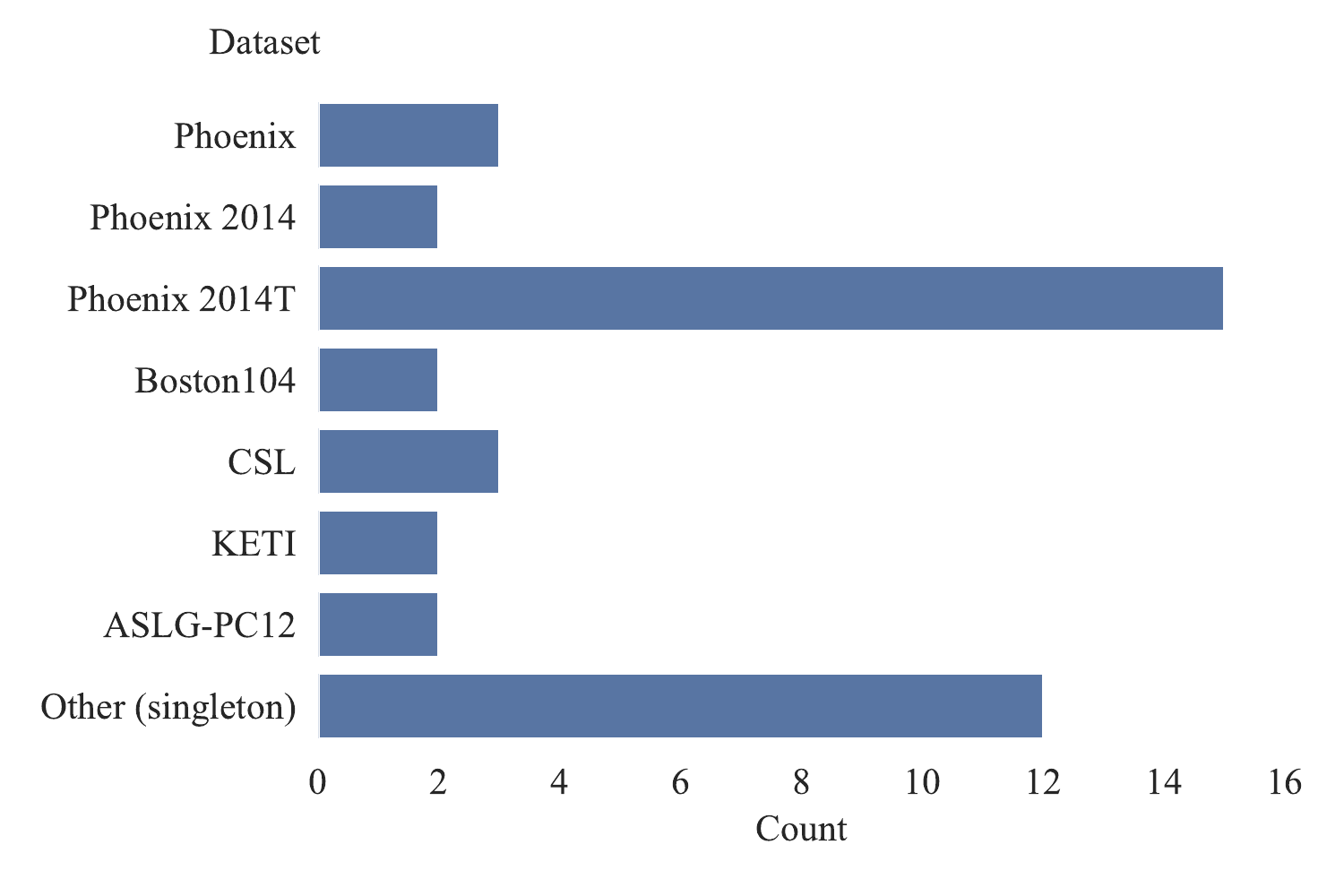}
\caption{The RWTH-PHOENIX-Weather 2014T dataset is used the most throughout literature. Other datasets are referenced at most three times in the 32 discussed papers.
Any dataset that occurs only one time is listed under ``Other (singleton)''.}
\label{fig:datasets}
\end{figure}

Several datasets are used in \ac{slt} research. Some are reused often, whereas others are only used once.
The distribution is shown in Fig.~\ref{fig:datasets}. It is clear that the most used dataset is
RWTH-PHOENIX-Weather 2014T \cite{camgoz2018neural} (used in 37\% of the reported papers).
This is because it is the first dataset large enough for neural \ac{slt} and because it is readily
available for research purposes.
This dataset is an extension of earlier versions, RWTH-PHOENIX-Weather \cite{forster2012rwth}
and RWTH-PHOENIX-Weather 2014 \cite{forster2014extensions}, for \ac{nmt}.
It contains videos in \ac{dgs}, gloss level annotations, and text in German. Precisely because
of the popularity of this dataset, we can compare several approaches to \ac{slt}: see Section~\ref{sec:benchmark}.

Other datasets are also reused several times.
The CSL dataset \cite{guo2018hierarchical} contains videos in \ac{csl} and text in Chinese.
The KETI dataset \cite{ko2019neural} contains \ac{ksl} videos, gloss level annotations, and Korean text. RWTH-Boston-104 \cite{stein2007hand} is a dataset for \ac{asl} to English translation containing \ac{asl} videos,
gloss level annotations, and English text.
The ASLG-PC12 dataset \cite{othman2012english} contains \ac{asl} glosses and English text.

An overview of dataset sizes as well as vocabulary sizes is given in Table~\ref{tab:dataset_stats}.
Note that the largest dataset in terms of number of parallel sentences, ASLG-PC12, contains 827 thousand training
sentences.
For \ac{mt} between spoken languages, datasets typically contain several millions of sentences, for example
the Paracrawl corpus \cite{EsplGomis2019ParaCrawlWP}. Furthermore,
the largest dataset with video data (RWTH-PHOENIX-Weather 2014T) contains only 7,096 training sentences.
It is clear that compared to spoken language datasets, sign language datasets lack labeled data.
In other words, \ac{slt} is a \emph{low-resource} \ac{mt} task.

12 papers (37.5\%) use custom datasets that are not publicly available \cite{ko2019neural,guo2018hierarchical,guo2019hierarchical,kim2020robust,luqman2020machine,stein2012analysis,morrissey2007combining,rodriguez2021important,lopez2010spanish,bungeroth2004statistical,rodriguez2020understanding,moe2020unsupervised}, limiting further analysis of their results as they cannot be compared directly to other papers.

\begin{table}
\caption{Statistics of the datasets that are used in more than one paper.}
\label{tab:dataset_stats}
\centering
\setlength{\tabcolsep}{3pt}
\begin{tabular}{|p{80pt}|p{70pt}|p{45pt}|p{40pt}|p{30pt}|p{45pt}|}
\hline
Dataset  & Languages   & Sentences & Signs  & Vocab. & Singletons \\
\hline
Phoenix \cite{forster2012rwth}& DGS-German  & 3,118      & 25,449  & 768        & 248 \\
Phoenix 2014 \cite{forster2014extensions} & DGS-German  & 9,015      & 102,726  & 1,580       & 565    \\
Phoenix 2014T \cite{camgoz2018neural} & DGS-German  & 8,257      & 75,783  & 1,870       & 337 \\
Boston-104 \cite{stein2007hand}     & ASL-English & 201       & 888    & 168        & 27 \\
CSL      \cite{guo2018hierarchical}     & CSL-Chinese & 5,000      & -      & 179           & - \\
KETI   \cite{ko2019neural}       & KSL-Korean  & 14,672      & -      & 524        & - \\
ASLG-PC12 \cite{othman2012english}    & ASL-English & 87,709     & 913,579 & 22,255      & 6,133 \\
\hline
\end{tabular}
\end{table}

\subsection{Evaluation}
The majority of evaluation of the quality of \ac{slt} models is based on quantitative metrics.
Eight different metrics are used for quantitative evaluation across the 32 papers: BLEU, ROUGE, WER, TER, PER, CIDEr, METEOR, COMET and NIST. The BLEU metric is used most often, by 29 (91\%) papers. ROUGE is used by fifteen (47\%) papers. The WER is reported on by seven (22\%) papers, TER by five (16\%) and PER by six (19\%).
CIDEr and METEOR are used four (13\%) and seven (22\%) times respectively, and COMET and NIST are reported on
by one paper each.
It is clear that the BLEU metric is the most popular metric, followed by ROUGE, WER and METEOR.

To the best of our knowledge, none of the papers discussed in this overview contain evaluations by members of \acp{slc}. One paper does perform human evaluation, but only by hearing people: Luqman \emph{et al.} \cite{luqman2020machine} let native Arabic
speakers evaluate the model's output translations as ``not understandable'', ``somehow understandable'' or ``understandable''. They define a
translation as acceptable if it is ``somehow understandable'' or ``understandable''. 80\% of the reported translations
are rated as understandable, 12\% are somehow understandable and 8\% are not understandable: therefore 92\% of the translations
are acceptable according to the authors. They also report the BLEU and TER metrics. They
remark that these metrics cannot handle different word orders, whereas certain word orders are interchangeable
in Arabic.
This is a drawback of using $n$-gram based metrics. These metrics depend
on the presence of several reference translations per example in the dataset to account for the fact that there can be
multiple correct word orders and to account for synonyms. However, often only a single
reference translation is provided, for example in the RWTH-PHOENIX-Weather 2014T dataset.

\subsection{The RWTH-PHOENIX-Weather 2014T benchmark}
\label{sec:benchmark}
The popularity of the RWTH-PHOENIX-Weather 2014T dataset facilitates the comparison of different
\ac{slt} models on this dataset.
% In this section, we aim to discover whether certain sign language representations are more powerful
% than others, and whether certain \ac{nmt} architectures consistently outperform others.
As the BLEU metric is the most common in these papers, we compare models
based on their BLEU-4 score.
% Based on our findings on this dataset, we aim to draw some conclusions about approaches to \ac{slt},
% while keeping in mind that these findings may not hold for other sign languages or datasets.

An overview of Gloss2Text models is shown in Table~\ref{tab:p2014t_g2t}.
For Sign2Gloss2Text, we refer to Table~\ref{tab:p2014t_s2g2t}.

For Sign2(Gloss+Text) and Sign2Text,
we list the results in Tables~\ref{tab:p2014t_s2gt} and~\ref{tab:p2014t_s2t}, respectively.

\begin{table}
\caption{Performance of different models on RWTH-PHOENIX-Weather 2014T Gloss2Text translation.}
    \label{tab:p2014t_g2t}
\centering
\setlength{\tabcolsep}{3pt}
\begin{tabular}{|p{20pt}|p{30pt}|p{80pt}|p{40pt}|p{40pt}|p{40pt}|p{40pt}|}
\hline
Ref.  & Year        & Architecture                & BLEU-4 & ROUGE & METEOR & COMET \\
\hline
\cite{camgoz2018neural} & 2018  & RNN          & 19.26  & 45.45 & -      & -     \\
\cite{partaourides2020variational} & 2020 & RNN            & 17     & 41.5  & -      & -     \\
            & &         & 16.7   & 40.7  & -      & -     \\
            & &                & 17     & 43.1  & -      & -     \\
            & &                & 18.1   & 43.5  & -      & -     \\
            & &            & 17.8   & 42.8  & -      & -     \\
\cite{camgoz2020sign}  & 2020 & Transformer          & 24.54  & -     & -      & -     \\
\cite{yin2020better}      & 2020 & Transformer          & 23.32  & 46.58 & 44.85  & -     \\
            & &  & 24.9   & 48.51 & 46.25  & -     \\
\cite{moryossef-etal-2021-data}   & 2021     & Transformer             & 22.02  & -     & -      & 6.84  \\
            & &         & 23.35  & -     & -      & 13.65 \\
            & &        & 23.17  & -     & -      & 11.7  \\
\cite{zhang-duh-2021-approaching} & 2021 & Transformer          & 24.38  & -     & -      & - \\
\hline
\end{tabular}
\end{table}

\subsubsection{Sign language representations}
We observe a wide variety of feature extraction methods across the different papers. They range from conceptually simple, frame based feature extractors, to linguistically motivated systems.

Several papers use features extracted using a 2D CNN by first training
a \ac{cslr} model on RWTH-PHOENIX-Weather 2014 \cite{camgoz2018neural,camgoz2020multi,de-coster-etal-2021-frozen}.
These papers use the full frame as inputs to the feature extractor.

Other papers combine multiple input channels.
Yin \emph{et al.} \cite{yin2020better} use \ac{stmc} features, extracting images
of the face, hands and full frames as well as including estimated poses of the body. These features
are processed by a network which performs temporal processing, both on the intra- as the inter-cue level. Their models are the \ac{sota} of Sign2Gloss2Text translation (25.4 BLEU-4). A similar approach is taken by Zhou \emph{et al.} \cite{zhou2021spatial}, whose model obtains a BLEU-4 score of 23.65 on Sign2(Gloss+Text) translation (\ac{sota}).
Camgoz \emph{et al.} use mouth pattern cues, pose information
and hand shape information; by using this multi-cue representation, they are able to remove glosses from their model and achieve competitive performance compared to models that do use glosses \cite{camgoz2020multi}.
The \ac{sota} is defined by the use of multi-cue features for Sign2Gloss2Text
\cite{yin2020better} and Sign2(Gloss+Text) \cite{zhou2021spatial} translation.

Frame based feature representations result in long input sequences to the translation model. The length of these sequences can be reduced by considering short clips instead of frames. We find this approach in scientific literature in two forms: (i) by using a pretrained 3D CNN or (ii) by reducing the sequence length using temporal convolutions or \acp{rnn} that are trained jointly with the translation model.
The benchmarking study of Orbay \emph{et al.} \cite{orbay2020neural}
suggests that 2D \acp{cnn} provide more informative features than 3D \acp{cnn}, especially if
the 2D \ac{cnn} is pretrained on a related task such as human pose estimation or hand shape classification.
The fact that 3D CNNs are outperformed by 2D CNNs in that study does not mean that spatial feature extractors always outperform spatio-temporal feature extractors.
Zhou \emph{et al.} \cite{zhou2021improving} use 2D CNN features extracted from
full frames, which are then further using temporal convolutions, reducing the temporal feature
size by a factor 4. They call this approach \ac{tin}. They achieve near-\ac{sota} performance
on Sign2Gloss2Text translation (23.51 BLEU-4) and \ac{sota} performance on Sign2Text translation (24.32 BLEU-4).
Zheng \emph{et al.} \cite{zheng2020improved} propose using an unsupervised algorithm
called \ac{fsdc} to remove temporally redundant frames by comparing frames on the level of pixels.
The resulting features are then processed using a combination of temporal convolutions and
\acp{rnn}. Zheng \emph{et al.} compare
the different settings and their combination and find that these techniques can be used
to reduce the input size of the sign language features as well as to improve the translation performance to 10.66 BLEU-4. They do not achieve performance on par with the \ac{sota} due to their choice of an underpowered feature extractor: AlexNet \cite{krizhevsky2012imagenet} pretrained on ImageNet \cite{deng2009imagenet}. This feature extractor was also used by Camgoz \emph{et al.} \cite{camgoz2018neural}, whose model achieved a BLEU-4 score of 9.58. It has been superseded by the more advanced ones mentioned above.
Clearly, temporal processing of the sign language representation before translation can improve the performance of \ac{slt} models, if the temporal processing module is trained jointly with the translation model so that it can exploit information
on sign language translation (e.g., \cite{yin2020better,zhou2021improving,zheng2020improved}). Otherwise, the temporal processing results in a loss of information (e.g., \cite{orbay2020neural}).

In conclusion, \ac{slt} models can be improved by applying the two following procedures. Firstly, features can be extracted from multiple channels linked to sign language parameters (e.g., hand and face crops). Secondly, temporal processing methods that are trained jointly with the translation model can reduce the sequence length and improve performance.

\subsubsection{Neural architectures}
We now determine whether recurrent models or transformers perform best on this dataset. As this may be dependent on the used sign language representation and translation model, we perform an analysis for Gloss2Text, Sign2Gloss2Text, Sign2(Gloss+Text) and Sign2Text, separately.

Because all Gloss2Text models use the same sign language representation, i.e., glosses, we can directly
compare the performance of different encoder-decoder architectures.
Here, we see that transformer based models perform better ($23.67 \pm 0.998$)
than recurrent models ($17.64 \pm 0.955$).

\begin{table}
\caption{Performance of different models on RWTH-PHOENIX-Weather 2014T Sign2Gloss2Text translation.}
    \label{tab:p2014t_s2g2t}
    \centering
\setlength{\tabcolsep}{3pt}
\begin{tabular}{|p{20pt}|p{30pt}|p{80pt}|p{80pt}|p{40pt}|p{40pt}|p{40pt}|}
\hline
Ref. & Year     & Repr.     & Architecture                     & BLEU-4 & ROUGE & METEOR \\
\hline
\cite{camgoz2018neural}    & 2018 & Spatial & RNN            & 18.13  & 43.8  & -      \\
\cite{camgoz2020sign}      & 2020 & Spatial & Transformer               & 22.45  & -     & - \\
\cite{yin2020better}    & 2020 &  Spatio-temporal, multi-cue         & RNN            & 21.54  & 45.5  & 44.87  \\
          & &              &                & 21.75  & 45.66 & 44.84  \\
          & &              & Transformer               & 24     & 46.77 & 45.78  \\
          & &              &           & 25.4   & 48.78 & 47.6   \\
\cite{zhou2021improving} & 2021 & Spatio-temporal          & Transformer & 23.51  & 49.35 & -      \\
\hline
\end{tabular}
\end{table}

We cannot as easily compare the performance between recurrent models and transformer based models for Sign2Gloss2Text
translation, because different papers use different feature extractors.
There is however one work that compares both architectures.
Yin \emph{et al.} \cite{yin2020better}
achieve better performance with transformers ($24.7 \pm 0.699$)
than with recurrent models ($21.65 \pm 0.105$).

\begin{table}
\caption{Performance of different models on RWTH-PHOENIX-Weather 2014T Sign2(Gloss+Text) translation.}
    \label{tab:p2014t_s2gt}
    \centering
\setlength{\tabcolsep}{3pt}
\begin{tabular}{|p{20pt}|p{30pt}|p{80pt}|p{80pt}|p{40pt}|p{40pt}|}
%\begin{tabular}{llll|rr}
\hline
Ref. & Year   & Representation    & Architecture       & BLEU-4 & ROUGE \\
\hline
\cite{camgoz2020sign} & 2020 & Spatial & Transformer & 21.32  & -     \\
\cite{zhou2021spatial}   & 2021 &  Spatio-temporal, multi-cue       & RNN        & 23.65  & 46.65 \\
\cite{de-coster-etal-2021-frozen}     & 2021 & Spatial & Transformer    & 22.25  & -     \\
    &   &             &    & 21.16  & -     \\
    &   &             &     & 16.64  & -     \\
\hline
\end{tabular}
\end{table}

No such comparison is available for Sign2(Gloss+Text) translation. We can only note that the best performing
model is that of Zhou \emph{et al.} \cite{zhou2021spatial}. It is an \ac{lstm}
encoder-decoder using spatio-temporal multi-cue features. These features are similar to the ones
used by Yin \emph{et al.} \cite{yin2020better}, who obtained better results using transformers
than using \acp{rnn} for Sign2Gloss2Text translation. Because of the lack of comparative papers, we are
unable to draw any conclusions with regards to the difference in performance between transformers
and \acp{rnn} for Sign2(Gloss+Text) translation.

\begin{table}
\caption{Performance of different models on RWTH-PHOENIX-Weather 2014T Sign2Text translation.}
    \label{tab:p2014t_s2t}
    \centering
\setlength{\tabcolsep}{3pt}
\begin{tabular}{|p{20pt}|p{30pt}|p{80pt}|p{80pt}|p{40pt}|p{40pt}|}
\hline
Ref. & Year          & Repr.                  & Architecture                             & BLEU-4 & ROUGE \\
\hline
\cite{camgoz2018neural}        & 2018 & Spatial   & RNN                       & 9.58   & 31.8  \\
\cite{camgoz2020sign} & 2020     & Spatial   & Transformer                        & 20.17  &      - \\
\cite{camgoz2020multi} & 2020 & Spatial, multi-cue            & Transformer                     & 19.21  & 45.05 \\
              & &                                   &        & 18.51  & 43.57 \\
\cite{orbay2020neural}  & 2020 & Spatial & RNN                       & 9.4    & 29.41 \\
              & & &                                   & 9.33   & 29.62 \\
              & &  &                                    & 9.06   & 29.09 \\
              & & Spatio-temporal          &                                    & 8.76   & 29.74 \\
              & &           &                                    & 8.26   & 28.64 \\
              & &           &                                    & 8.09   & 28    \\
              & & Spatial (pose)       &                                    & 10.92  & 32.85 \\
              & &          &                                    & 10.23  & 31.47 \\
              & &          &                                    & 9.91   & 30.65 \\
              & & Spatial                  &                                    & 12.17  & 34.59 \\
              & &                  &                                    & 11.15  & 31.98 \\
              & &          &                                    & 12.21  & 34.41 \\
              & &  &                                    & 13.25  & 36.28 \\
\cite{zheng2020improved}    & 2020  & Spatial                    & RNN                            & 9.76   & 31.34 \\
              & &                           &          & 12.4   & 31.2  \\
              & &                           &  & 10.66  & 32.25 \\
              & &                           &          & 9.71   & 31.52 \\
              & &                           &                   & 10.73  & 32.99 \\
\cite{zhao2021conditional}   & 2021 & Spatial                   & Transformer                    & 15.18  & 38.85 \\
\cite{rodriguez2021important} & 2021 & Spatio-temporal              & RNN                       & 4.56   & -      \\
\cite{zhou2021improving}     & 2021 & Spatio-temporal  & Transformer      & 24.32  & 49.54 \\
\hline
\end{tabular}
\end{table}

The Sign2Text translation models exhibit higher variance in their scores than models for the other tasks. This is due to the lack of additional supervision signal in the form of glosses: the choice of sign language representation has a larger impact on the translation score.
The difference in BLEU-4 score between transformers ($19.478 \pm 3.294$) and
\acp{rnn} ($10.007 \pm 1.905$) is larger than in other tasks. This is possibly because transformers
are better able to handle long term dependencies through the attention mechanism: without glosses,
the dependencies in the source language sentences can be quite long.

\begin{figure}
\centering
\includegraphics[width=0.45\textwidth]{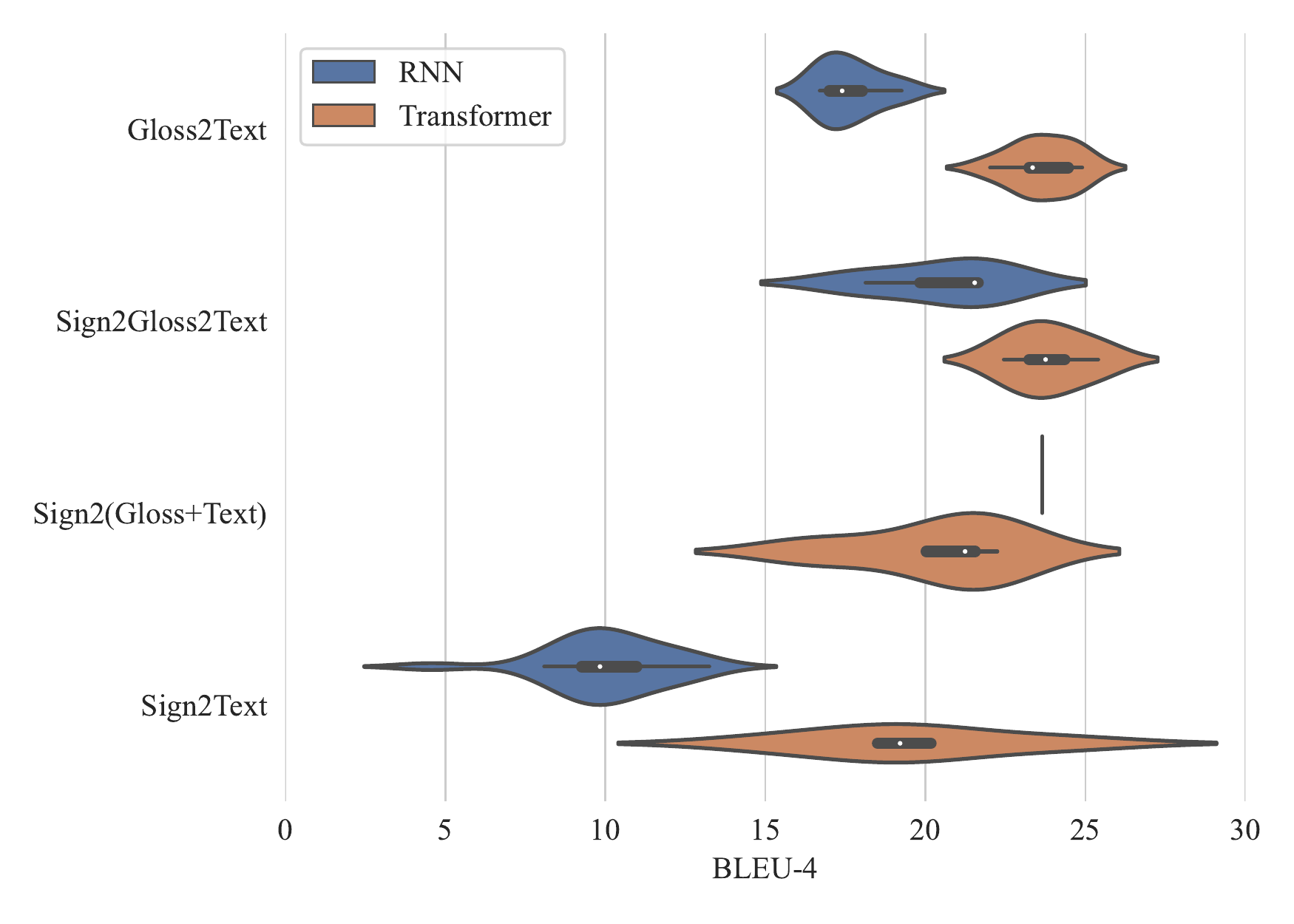}
\caption{Transformers tend to outperform \acp{rnn} on different \ac{slt} tasks in terms of BLEU-4 score on the RWTH-PHOENIX-Weather 2014T dataset.}
\label{fig:rnn_vs_transformer}
\end{figure}

We provide a graphical overview of the performance of \acp{rnn} and transformers across tasks in Fig.~\ref{fig:rnn_vs_transformer}
and conclude that transformers outperform \acp{rnn} in most cases on RWTH-PHOENIX-Weather 2014T.

\section{Discussion of the current state of the art}
\label{sec:discussion}
%\subsection{Summary}
By analyzing the existing scientific literature on \ac{slt}, we set out to discover answers to the following questions.
Which kinds of sign language representations are most informative (\ref{rq:sl_representation})?
Which types of models should be used for \ac{slt} (\ref{rq:slt_sota})?
Which datasets are currently used (\ref{rq:datasets})? How can we evaluate \ac{slt} models (\ref{rq:evaluation})?
Because of the diversity in applied methods and datasets, answering these questions definitively
is challenging. However, the popularity of the RWTH-PHOENIX-Weather 2014T benchmark has enabled
some comparisons between different models and provides us with several insights.

\subsection{Sign language representations}\label{sec:discussions_sl_representations}
\ref{rq:sl_representation} asks, \emph{``Which kinds of sign language representations are most informative?''}
The best performing \ac{slt} models all use either (i) deep learning-based sign language representations or (ii) glosses. Regarding (i), most commonly these representations are extracted using \acp{cnn} that are pretrained on related tasks (e.g., hand shape classification or \ac{cslr}).

Several feature extraction techniques are used in the literature. They range from 2D \acp{cnn} to human pose estimation and 3D \acp{cnn}, and combinations thereof.
Translation models benefit from sign language representations extracted from multiple cues (e.g., hand crops, face crops and pose estimation). Human pose estimation is primarily useful as such a cue rather than as the only feature extractor \cite{camgoz2020multi,yin2020better}: current techniques often fail in the presence of motion blur and manual-facial interactions which are common in sign language data \cite{de2020sign,moryossef2021evaluating}.
After the sign language representation (i.e., a sequence of such features) has been extracted, its length can be reduced to further improve the translation performance. This step must be trained end-to-end with the translation model \cite{yin2020better,zhou2021improving,zheng2020improved}. Otherwise, if the sequence is simply reduced at arbitrary steps (as for example a frozen pretrained 3D \ac{cnn} would do), the translation model's performance suffers \cite{orbay2020neural}.

Earlier papers using \ac{smt} methods often include linguistic properties of sign languages in their models as a way to
reduce the problem complexity, for example by computing features that represent hand trajectories. In contrast,
many recent \ac{nmt}-based papers prefer to use end-to-end deep learning. Given the low-resource nature of current
\ac{slt} datasets (often only several thousands of parallel sentences), the incorporation of domain knowledge
(in this case, linguistic knowledge) proves beneficial to the quality of the translations. 

Glosses (ii), as written sign language representations which aim to capture the meaning of signs, are easily adopted in sequence-to-sequence models, as both written text and signs can be mapped to glosses. Out of the 32 papers we reviewed, 23 use glosses in some way.
Unfortunately, using glosses has several drawbacks. Firstly, annotating sign language corpora on the gloss level
is a time-consuming process typically performed by domain experts.
Secondly, any translation model that uses glosses as a sign language
representation (i.e., Gloss2Text and Sign2Gloss2Text models), requires
a sign language recognition system of sufficient quality also at
inference time. Finally, glosses do not accurately capture the entire
meaning of signs.
Some research attempts to remove the gloss dependency,
instead proposing so-called Sign2Text systems which translate directly from sign language videos into
spoken language texts.
Sign2Text systems are slowly becoming
competitive with systems dependent on glosses in terms of translation scores such as BLEU-4. A large part
of this is due to the incorporation of domain knowledge in the design
of the sign language representations, for example including mouth
patterns as an additional information channel \cite{yin2020better,zhou2021spatial,camgoz2020multi}.

\subsection{Translation model architectures}
Once a proper sign language representation is determined, a translation model must be designed. \ref{rq:slt_sota} asks whether there is one superior algorithm for \ac{slt}.
Despite the generally small size of the datasets used for \ac{slt}, we see that neural \ac{mt} models
achieve the highest translation scores. In particular, transformers appear to outperform \acp{rnn}
in several cases --- but not consistently: it depends on the task and the used sign language representation.
The fact that pretrained language models are readily available for many transformer based architectures (for example via the HuggingFace Transformers library \cite{Wolf_Transformers_StateoftheArt_Natural_2020}) may give
transformers an edge over \acp{rnn}. De Coster \emph{et al.} \cite{de-coster-etal-2021-frozen} have
shown that integrating pretrained spoken language models in an \ac{slt} model can increase the BLEU-4 score compared
to training transformers from scratch on \ac{slt}. Furthermore, the attention mechanism in
transformers can be used to inspect the model's decision making \cite{vaswani2017attention}. This
was for example previously performed for isolated \ac{slr}, showing that transformers focus on
distinguishing frames in clips \cite{de2021isolated}. To the best of our knowledge, no such analysis has been published yet
for transformers in \ac{slt}. It may prove useful for error analysis of current translation models.

\subsection{Datasets}
Neural translation models need to be trained with sufficiently large datasets. The third question we set out to answer, \ref{rq:datasets}, is, \emph{``Which datasets are used and what are their properties?''}
The most used dataset is RWTH-PHOENIX-Weather 2014T \cite{camgoz2018neural} for translation from \ac{dgs} to German.
It contains 8,257 parallel utterances from several different interpreters. The domain is weather broadcasts. The fact that it is used so often, allows
for comparisons between different methods and allows for incremental progress in \ac{slt}.

Several papers use custom datasets, and often only once.
Custom datasets can be useful to validate existing approaches
on different languages, but the majority of progress is made on public benchmark datasets such as
RWTH-PHOENIX-Weather 2014T because such datasets enable the comparison of different approaches.
In contrast, using private datasets limits the reproducibility of the presented models. Preferably,
a model is trained and validated on a benchmark and then one can evaluate the same model on private datasets, perhaps in a local sign language.

Current datasets have several limitations. They are typically restricted to limited domains of discourse (weather broadcasts) and have little variability in terms of visual conditions (TV studios). Furthermore, some sign language translation datasets contain recordings of nonnative signers. In cases, the signing is interpreted (under time pressure) from spoken language. This means that the used signing may not be representative of the sign language and may in fact be influenced by the grammar of a spoken language. Training a translation model on these kinds of data has implications for the quality and accuracy of the resulting translations.

\subsection{Tasks}
A training scheme is needed to train these translation models on the proposed data. In literature, we have found four such schemes: Gloss2Text, Sign2Gloss2Text, Sign2(Gloss+Text) and Sign2Text. Two of these require gloss annotations for all examples (Gloss2Text and Sign2Gloss2Text). Sign2(Gloss+Text) requires gloss annotations for some or all examples (and does not require glosses at inference time) and Sign2Text does not use glosses at all.

\begin{figure}
\centering
\includegraphics[width=\textwidth]{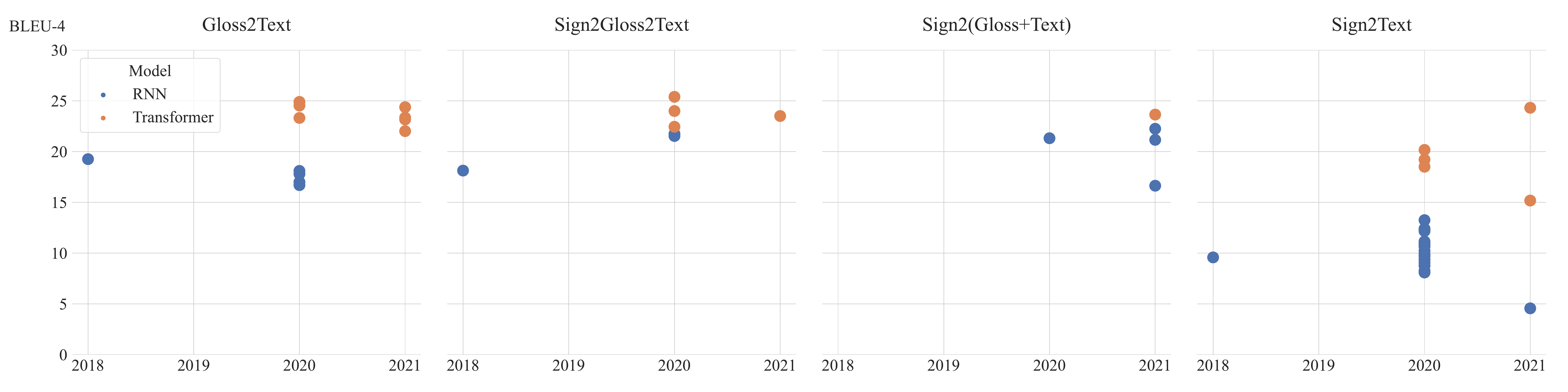}
\caption{We can observe a general trend towards higher BLEU-4 scores on the RWTH-PHOENIX-Weather 2014T
dataset for all tasks. As recently as 2021,
there is little difference between the top scores of different tasks.}
\label{fig:phoenix}
\end{figure}

Fig.~\ref{fig:phoenix} shows an overview of the BLEU-4 scores on the RWTH-PHOENIX-Weather 2014T dataset
from its release in 2018 until August 2021. We clearly see increases in scores for all four tasks,
though the increase has not yet continued into 2021 for Gloss2Text and Sign2Gloss2Text.
We see a rising trend for Sign2(Gloss+Text) and Sign2Text, as there is no
gloss bottleneck and improvements in feature extraction techniques
as well as architecture enable better translation scores.
In most cases, transformers outperform \acp{rnn} based methods on RWTH-PHOENIX-Weather 2014T.

\subsection{Evaluation}
In terms of evaluation, we see many papers reporting several translation related metrics, such as BLEU,
ROUGE, WER and METEOR. These are standard metrics in \ac{mt}. Several papers also provide
example translations to allow the reader to gauge the translation quality for themselves.
Whereas the above metrics
often correlate quite well with human evaluation, this is not always the case \cite{callison2006re}.
Including human evaluation is especially
important for spoken to signed language translation, where avatars must sign in a natural way. However, it is also paramount for signed to spoken language translation to assess the fluency and correctness of
translations. Only one of the 32 reviewed papers
incorporates human evaluators in the loop \cite{luqman2020machine}.
In none of the reviewed papers is evaluation in collaboration with \ac{slc} members mentioned. As the translation models proposed in these papers are designed first and foremost for communication between nonsigners and sign language users, they should all be involved in the evaluation of those models.

\section{Challenges and proposals}
\label{sec:challenges}

\subsection{Sign language representations}
\ac{slt} models require salient representations of sign languages as input, as noted in Section~\ref{sec:discussions_sl_representations}. The \ac{slt} models we investigated are primarily adapted spoken language MT models. Therefore, they expect sign language representations to contain information on the meaning of signs, similar to the function of word embeddings in spoken language \ac{mt}.
There is currently no representation that contains all of the information required for proper sign language translation and that takes into account both
the established \emph{and} the productive lexicons.
In fact, it is doubtful whether a pure end-to-end \ac{nmt} approach is capable of tackling productive signs. To recognize and understand productive signs, we need models that have the ability to link abstract visual information to the properties of objects. This relates to caption generation models and models trained to classify, e.g., highly abstracted drawings. Incorporating the productive lexicon in translation systems is a significant challenge, one for which, to the best of our knowledge, labeled data are currently not available.

The majority of \ac{slt} models, especially neural models, overlook several linguistic elements of
signing. Especially in recent years, \ac{slt} is tackled using end-to-end deep learning techniques. The used representations are typically phonological, focusing on detecting hand shapes and movements.
These representations form a good basis for a translation model, but they do not contain meaning. They also do not model fingerspelling, signing space, or classifiers explicitly. Learning these aspects in the translation model with an end-to-end approach is challenging.
This is made even more difficult by the lack of annotated data.

The design of a salient sign language representation relates to the problem of sign language segmentation. De Sisto \emph{et al.} \cite{de-sisto-etal-2021-defining} describe why segmentation into individual signs is difficult, citing among others
coarticulation and simultaneity.
The definition and extraction of so-called ``meaningful units'' for \ac{slt} is
an open research question that needs to be answered in order to move beyond frame based and clip based phonological representations. Collaboration between computer scientists and (computational) linguists can aid
in this effort.

\subsection{Exploiting the vast amounts of unlabeled data}
Current \ac{sota} \ac{slt} models use sign language representations created through supervised end-to-end deep learning, i.e., on labeled datasets.
The collection and annotation process of these datasets is expensive and time intensive. However, there are already vast amounts of
unannotated sign language videos. This raises the question how
unsupervised machine learning techniques can be exploited in the domain of \ac{slt}.

In the domain of natural language processing, we already observe tremendous advances thanks
to unsupervised language models such as BERT \cite{devlin2018bert}. Whereas such (pretrained) models have been exploited
in \ac{slt} papers \cite{de-coster-etal-2021-frozen,zhao2021conditional}, they have not yet been
trained specifically for the purpose of processing sign language videos.

In computer vision, self-supervised techniques are applied to pretrain powerful
feature extractors which can then be applied to downstream tasks such as image classification
or object detection. Algorithms such as SimCLR \cite{chen2020simple},
BYOL \cite{grill2020bootstrap} and DINO \cite{caron2021emerging}
are used to train 2D \acp{cnn} without labels, reaching performance that is almost on the same
level as models trained with supervised techniques.

Up to a certain level, sign languages
share some common elements, for example the fact that they all use the human body to convey information.
Movements used in signing are composed of motion primitives and
the configuration of the hand (shape and orientation) is important in all sign languages.
The recognition of these low level components does not require
language specific datasets and could be performed on multilingual datasets, containing videos
recorded around the world from people with various ages, genders, and ethnicities.
A parallel can be drawn to speech recognition:
Wav2Vec 2.0, for example, learns discrete speech units in
a self-supervised manner \cite{baevski2020wav2vec}; a similar approach could be beneficial for
pretraining on unlabeled sign language videos.

Deep neural networks trained in a self-supervised way could be applied to learn such common elements of different sign languages. This would not only facilitate automatic \ac{slt}, but could also lead to the development of new tools supporting linguistic analysis of sign languages.
However, there has been very little investigation into these matters in the
scientific literature. Given that annotated data is so scarce in this domain,
we advocate for the investigation of unsupervised machine learning techniques.

\subsection{Dataset collection}
Currently, \ac{slt} from sign language videos to spoken language text is a low-resource \ac{mt} task,
with the largest public dataset containing just 7,096 training examples \cite{camgoz2018neural}. Furthermore, far from all sign languages have corresponding translation datasets.
Additional datasets need to be collected and existing ones need to be extended. Such collection efforts are expensive, and there are several
caveats.

De Meulder
\cite{de-meulder-2021-good} raises concerns with current dataset collection efforts.
Existing datasets and those currently being collected suffer from several biases. If interpreted
data are used, influence from spoken languages will be present in the dataset. If only native signer
data are used, then the majority of signers will have the same ethnicity. Both statistical as well as neural MT exacerbate bias \cite{vanmassenhove-etal-2019-lost,vanmassenhove-2021-MachineTranslationese}. Therefore, when our training datasets are biased and of small volumes, we cannot expect (data driven) \ac{mt} systems to reach high qualities and be generalizable.

We remark on the need for two kinds of datasets. On the one hand, sign language recognition, for the purpose of extracting sign language representations, requires large and varied datasets. These can be collected as part of new efforts, or collated from existing ones. If unsupervised techniques are applied (as mentioned above), then these datasets do not necessarily need to be labeled entirely. They also do not need to consist entirely of native signing. On the other hand, sign language translation requires high quality labeled data, the collection of which is challenging.
Bragg \emph{et al.}'s first and second calls to action, ``Involve Deaf team members throughout'' and ``Focus on real-world applications'' \cite{bragg2019sign}, can be a guide
during the dataset collection process.
By involving \ac{slc} members, the dataset collection effort can be guided towards use cases
that would benefit \acp{slc}.
Additionally, by collecting datasets with a limited domain of discourse targeted at specific use cases,
the \ac{slt} problem is effectively simplified. As a result, any applications would be limited in scope, but more useful in practice.

Current research uses datasets in which the videos have fixed viewpoints, similar backgrounds, and sometimes even, the signers wear similar clothing for maximum contrast with the background. In real-world applications, dynamic viewpoints and lighting conditions will be a common occurrence. Dataset collection efforts need to take this into account.

Sincan \emph{et al.}'s AUTSL dataset addresses these concerns, but for sign language recognition \cite{sincan2020autsl}. They focused on gathering a large dataset with variability in participants and their knowledge of sign language (including native and nonnative signers). Videos were recorded from different viewpoints and in different locations. Similar dataset collection efforts can aid the domain of \ac{slt}, reducing the discussed forms of bias and improving the translation quality for specific use cases.

\subsection{Computational performance}
Within the reviewed papers, there is little to no discussion on the computational performance of the proposed
methods. If the end goal is real time \ac{slt}, researchers need to be concerned with the types
of approaches used. Of course, achieving acceptable translation quality takes priority over
computational performance in the current stage of \ac{slt} research.

Research into efficient deep learning is a hot topic, as deep neural networks are now deployed on mobile and embedded devices.
Engineering techniques such as network quantization can reduce the computational cost of deep
neural networks \cite{gholami2021survey}.
At the same time, theoretical properties of neural networks are being investigated,
for example to reduce the computational complexity of transformers \cite{wang2020linformer}
and 2D \acp{cnn} \cite{tan2019efficientnet}. \ac{slt} models will be able
to benefit from these optimizations in the future.

Therefore, rather than optimizing translation models at this stage in research, we propose to simply keep computational efficiency in mind but focus first on translation quality. When translation quality reaches levels that are acceptable for end users, then the advances in deep learning can be exploited to make these existing models more efficient.

In the context of computational performance and resource consumption we cannot ignore the environmental impact of deep neural models. Whereas Strubell \emph{et al.}~\cite{strubell-etal-2019-energy} and Schwartz \emph{et al.}~\cite{Schwartz2020GreenA} advocate for greener \ac{nmt}, consistently larger models are still being trained. Perhaps when deciding our the architecture of an \ac{slt} model to be trained, we should pay more attention the fact that we are dealing with a low-resource problem and as such adapt design our systems adequately --- consider for example the paper of Sennrich and Zhang \cite{sennrich-zhang-2019-revisiting}. More compact models, trained faster on smaller, but use case specific datasets can be a computationally more efficient, more ecological and (use case specific) more effective option.

\subsection{Evaluation}
Current research uses mostly quantitative metrics to evaluate \ac{slt} models, on relatively simple datasets. Models should also be evaluated on real-world data from real-world settings. Furthermore, human evaluation from signers and nonsigners is required to truly assess the translation quality.

Human-in-the-loop development can not only yield better end results, but also alleviate some of the concerns that live in \acp{slc} about the application of \ac{mt} techniques to sign languages about appropriation and alteration of sign languages. Instead of performing research in isolation and then presenting a result to sign language users (the \emph{waterfall} technique), an \emph{agile} approach should be used. This could prove especially beneficial in the domain of \ac{slt} specifically: many researchers are hearing and do not use signing as their primary means of communication. By frequently evaluating models together with sign language users, we as a research community can avoid drifting off course towards unusable models. Wolfe \emph{et al.} state with regards to sign language avatars: ``the quality of the ultimate signed language display must be given highest priority in a spoken to signed translation system.''~\cite{wolfe2021myth} We argue that the same holds for translation from signed to spoken languages.

Finally, we see that in depth error analysis is missing from many \ac{slt} papers.
By (visually and/or numerically) analyzing
which types of errors are made by our models, we can make them more robust and focus our improvement efforts to where they are most needed. This will benefit the applicability
of researched techniques in real-world settings.

\subsection{The need for interdisciplinary research}
Sign language \ac{mt} cannot be tackled by computer scientists or linguists alone. There is a need
for collaboration between technical and linguistic profiles, driven by the input and augmented by the feedback of stakeholders,
namely the members of \acp{slc}. Without linguists, computer scientists are prone to missing
crucial aspects of sign language grammars. Without the watchful eye of \acp{slc},
researchers may lose track of which types of applications are desired by end users.
Computer scientists, finally, are needed, because
\ac{slt} is a highly challenging problem from a technical point of view, both in terms
of machine learning and in terms of developing real time applications. Only through the
consistent and close-knit collaboration of these groups, can an \ac{slt} system
that matches the needs of \ac{slc} members ever be developed.

Within the European Union, two international research projects have been launched as recently
as January 2021: SignON \cite{shterionov2021signon} and EASIER \cite{easier}.
The consortia in both of these projects are composed of computer scientists, linguists
and representatives of \acp{slc}. The interdisciplinary collaborations within
these projects have the potential to accelerate sign language translation research in the direction
of applications that are useful for the end users.

\section{Conclusion}
\label{sec:conclusion}
In this article, we discuss the \ac{sota} of \ac{slt} and explore challenges and opportunities for future research through a systematic overview of the papers in this domain.
We review 32 papers on sign language machine translation from a signed to a spoken language. These papers are selected based on predefined criteria and they are indicative of sound \ac{slt} research. The selected papers are written in English and peer reviewed. They propose, implement and
evaluate a sign language machine translation system from a sign language to a spoken language using an RGB
camera and do not contain descriptions that are explicitly offensive to members of the sign language communities. We discuss the \ac{sota} of \ac{slt} and explore several challenges and opportunities for future research.

In recent years, neural machine translation has become dominant in the growing domain of \ac{slt}. The most powerful sign language representations are those that combine information from multiple channels (manual actions, body movements and mouth patterns) and those that are reduced in length by temporal processing modules trained jointly with the translation model. These translation models are typically \acp{rnn} or transformers. Transformers outperform \acp{rnn} in many cases. \ac{slt} datasets are small: we are dealing with a low-resource machine translation problem. The datasets consider limited domains of discourse and generally contain recordings of nonnative signers. This has implications on the quality and accuracy of translations generated by models trained on these datasets, which must be taken into account when evaluating \ac{slt} models.
This evaluation is mostly performed using quantitative metrics that can be computed automatically, given a corpus. There are currently no papers that perform evaluation in collaboration with sign language users.

Progressing beyond the current \ac{sota} of \ac{slt} requires efforts in data collection, the design of sign language representations, machine translation, and evaluation.
Future research may improve sign language representations by incorporating domain knowledge into their design and by leveraging abundant, but as of yet unexploited, unlabeled data. Research should be conducted in an interdisciplinary manner, with computer scientists, sign language linguists, and experts on sign language cultures working together. Finally, sign language translation models should be evaluated in collaboration with end users: native signers as well as hearing people that do not know any sign language.

\section*{Acknowledgment}

M.  De  Coster's  research  is  funded  by  the  Research Foundation Flanders (FWO Vlaanderen):  file number 77410. This work has been conducted within the SignON project. This project has received funding from the European Union's Horizon 2020 research and innovation programme under grant agreement No 101017255.
M. De Coster thanks F. wyffels and P. Wolfert for their comments and suggestions.

\bibliographystyle{IEEEtran}
\bibliography{article}

\begin{thebibliography}{10}
\providecommand{\url}[1]{#1}
\csname url@samestyle\endcsname
\providecommand{\newblock}{\relax}
\providecommand{\bibinfo}[2]{#2}
\providecommand{\BIBentrySTDinterwordspacing}{\spaceskip=0pt\relax}
\providecommand{\BIBentryALTinterwordstretchfactor}{4}
\providecommand{\BIBentryALTinterwordspacing}{\spaceskip=\fontdimen2\font plus
\BIBentryALTinterwordstretchfactor\fontdimen3\font minus
  \fontdimen4\font\relax}
\providecommand{\BIBforeignlanguage}[2]{{%
\expandafter\ifx\csname l@#1\endcsname\relax
\typeout{** WARNING: IEEEtran.bst: No hyphenation pattern has been}%
\typeout{** loaded for the language `#1'. Using the pattern for}%
\typeout{** the default language instead.}%
\else
\language=\csname l@#1\endcsname
\fi
#2}}
\providecommand{\BIBdecl}{\relax}
\BIBdecl

\bibitem{erard2017sign}
\BIBentryALTinterwordspacing
M.~Erard, ``Why sign language gloves don't help deaf people,'' 2017. [Online].
  Available:
  \url{https://www.theatlantic.com/technology/archive/2017/11/why-sign-language-gloves-dont-help-deaf-people/545441/}
\BIBentrySTDinterwordspacing

\bibitem{vermeerbergen201336}
M.~Vermeerbergen, J.~N. Twilhaar, and M.~Van~Herreweghe, ``Variation between
  and within sign language of the netherlands and flemish sign language,'' in
  \emph{Language and Space Volume 30 (3): Dutch}.\hskip 1em plus 0.5em minus
  0.4em\relax Berlin: De Gruyter Mouton, 2013, pp. 680--699.

\bibitem{van2009flemish}
M.~Van~Herreweghe and M.~Vermeerbergen, ``Flemish sign language
  standardisation,'' \emph{Current issues in language planning}, vol.~10,
  no.~3, pp. 308--326, 2009.

\bibitem{stokoe1960sign}
W.~Stokoe, ``Sign language structure: An outline of the visual communication
  systems of the american deaf,'' \emph{Studies in Linguistics, Occasional
  Papers}, vol.~8, 1960.

\bibitem{battison1978lexical}
R.~Battison, \emph{Lexical borrowing in American sign language.}\hskip 1em plus
  0.5em minus 0.4em\relax Silver Spring: Linstok Press, 1978.

\bibitem{bank2011variation}
R.~Bank, O.~A. Crasborn, and R.~Van~Hout, ``{Variation in mouth actions with
  manual signs in Sign Language of the Netherlands (NGT)},'' \emph{Sign
  Language \& Linguistics}, vol.~14, no.~2, pp. 248--270, 2011.

\bibitem{perniss201219}
P.~Perniss, ``19. use of sign space,'' in \emph{Sign Language}.\hskip 1em plus
  0.5em minus 0.4em\relax De Gruyter Mouton, 2012, pp. 412--431.

\bibitem{zwitserlood2012classifiers}
\BIBentryALTinterwordspacing
I.~Zwitserlood, \emph{Classifiers}.\hskip 1em plus 0.5em minus 0.4em\relax De
  Gruyter Mouton, 2012, pp. 158--186. [Online]. Available:
  \url{https://doi.org/10.1515/9783110261325.158}
\BIBentrySTDinterwordspacing

\bibitem{vermeerbergen2006past}
M.~Vermeerbergen, ``Past and current trends in sign language research,''
  \emph{Language \& Communication}, vol.~26, no.~2, pp. 168--192, 2006.

\bibitem{FrishbergHoitingSlobin_SignTranscription}
\BIBentryALTinterwordspacing
N.~Frishberg, N.~Hoiting, and D.~I. Slobin, \emph{Transcription}.\hskip 1em
  plus 0.5em minus 0.4em\relax De Gruyter Mouton, 2012, pp. 1045--1075.
  [Online]. Available: \url{https://doi.org/10.1515/9783110261325.1045}
\BIBentrySTDinterwordspacing

\bibitem{sutton1981sign}
V.~Sutton, \emph{Sign writing for everyday use}.\hskip 1em plus 0.5em minus
  0.4em\relax Sutton Movement Writing Press, 1981.

\bibitem{Prillwitz1989hamnosys}
\BIBentryALTinterwordspacing
S.~Prillwitz, \emph{HamNoSys Version 2.0. Hamburg Notation System for Sign
  Languages: An Introductory Guide}, ser. Intern. Arb. z. Geb{\"a}rdensprache
  u. Kommunik.\hskip 1em plus 0.5em minus 0.4em\relax Signum Press, 1989.
  [Online]. Available: \url{https://books.google.nl/books?id=4DlAOgAACAAJ}
\BIBentrySTDinterwordspacing

\bibitem{vermeerbergen2007simultaneity}
M.~Vermeerbergen, L.~Leeson, and O.~A. Crasborn, \emph{Simultaneity in signed
  languages: Form and function}.\hskip 1em plus 0.5em minus 0.4em\relax John
  Benjamins Publishing, 2007, vol. 281.

\bibitem{Sign_A_Thesis}
I.~E. Murtagh, ``A linguistically motivated computational framework for irish
  sign language,'' Ph.D. dissertation, Trinity College Dublin.School of
  Linguistic Speech and Comm Sci, 2019.

\bibitem{de-sisto-etal-2021-defining}
\BIBentryALTinterwordspacing
M.~De~Sisto, D.~Shterionov, I.~Murtagh, M.~Vermeerbergen, and L.~Leeson,
  ``Defining meaningful units. challenges in sign segmentation and
  segment-meaning mapping,'' in \emph{Proceedings of the 1st International
  Workshop on Automatic Translation for Signed and Spoken Languages
  (AT4SSL)}.\hskip 1em plus 0.5em minus 0.4em\relax Virtual: Association for
  Machine Translation in the Americas, Aug. 2021, pp. 98--103. [Online].
  Available: \url{https://aclanthology.org/2021.mtsummit-at4ssl.11}
\BIBentrySTDinterwordspacing

\bibitem{sutskever2014sequence}
I.~Sutskever, O.~Vinyals, and Q.~V. Le, ``Sequence to sequence learning with
  neural networks,'' in \emph{Advances in neural information processing
  systems}, 2014, pp. 3104--3112.

\bibitem{Hochreiter1997LSTM}
\BIBentryALTinterwordspacing
S.~Hochreiter and J.~Schmidhuber, ``{Long Short-Term Memory},'' \emph{Neural
  Computation}, vol.~9, no.~8, pp. 1735--1780, 11 1997. [Online]. Available:
  \url{https://doi.org/10.1162/neco.1997.9.8.1735}
\BIBentrySTDinterwordspacing

\bibitem{cho2014neural}
K.~Cho, B.~van Merri{\"{e}}nboer, {\c C}.~G{\"{u}}l{\c c}ehre, D.~Bahdanau,
  F.~Bougares, H.~Schwenk, and Y.~Bengio, ``Learning phrase representations
  using {RNN} encoder--decoder for statistical machine translation,'' in
  \emph{Proceedings of the 2014 Conference on Empirical Methods in Natural
  Language Processing}, Doha, Qatar, 2014, pp. 1724--1734.

\bibitem{bahdanau2015neural}
D.~Bahdanau, K.~H. Cho, and Y.~Bengio, ``Neural machine translation by jointly
  learning to align and translate,'' in \emph{3rd International Conference on
  Learning Representations, ICLR 2015}, 2015.

\bibitem{vaswani2017attention}
A.~Vaswani, N.~Shazeer, N.~Parmar, J.~Uszkoreit, L.~Jones, A.~N. Gomez,
  {\L}.~Kaiser, and I.~Polosukhin, ``Attention is all you need,'' in
  \emph{Advances in neural information processing systems}, 2017, pp.
  5998--6008.

\bibitem{pennington2014glove}
\BIBentryALTinterwordspacing
J.~Pennington, R.~Socher, and C.~D. Manning, ``Glove: Global vectors for word
  representation,'' in \emph{Empirical Methods in Natural Language Processing
  (EMNLP)}, 2014, pp. 1532--1543. [Online]. Available:
  \url{http://www.aclweb.org/anthology/D14-1162}
\BIBentrySTDinterwordspacing

\bibitem{devlin2018bert}
\BIBentryALTinterwordspacing
J.~Devlin, M.-W. Chang, K.~Lee, and K.~Toutanova, ``{BERT}: Pre-training of
  deep bidirectional transformers for language understanding,'' in
  \emph{Proceedings of the 2019 Conference of the North {A}merican Chapter of
  the Association for Computational Linguistics: Human Language Technologies,
  Volume 1 (Long and Short Papers)}.\hskip 1em plus 0.5em minus 0.4em\relax
  Minneapolis, Minnesota: Association for Computational Linguistics, Jun. 2019,
  pp. 4171--4186. [Online]. Available:
  \url{https://www.aclweb.org/anthology/N19-1423}
\BIBentrySTDinterwordspacing

\bibitem{Lewis2020Bart}
\BIBentryALTinterwordspacing
M.~Lewis, Y.~Liu, N.~Goyal, M.~Ghazvininejad, A.~Mohamed, O.~Levy, V.~Stoyanov,
  and L.~Zettlemoyer, ``{BART:} denoising sequence-to-sequence pre-training for
  natural language generation, translation, and comprehension,'' in
  \emph{Proceedings of the 58th Annual Meeting of the Association for
  Computational Linguistics, {ACL} 2020, Online, July 5-10, 2020}, D.~Jurafsky,
  J.~Chai, N.~Schluter, and J.~R. Tetreault, Eds.\hskip 1em plus 0.5em minus
  0.4em\relax Association for Computational Linguistics, 2020, pp. 7871--7880.
  [Online]. Available: \url{https://doi.org/10.18653/v1/2020.acl-main.703}
\BIBentrySTDinterwordspacing

\bibitem{Harris1954_DistributionalStructure}
\BIBentryALTinterwordspacing
Z.~Harris, ``Distributional structure,'' \emph{Word}, vol.~10, no. 2-3, pp.
  146--162, 1954. [Online]. Available:
  \url{https://link.springer.com/chapter/10.1007/978-94-009-8467-7_1}
\BIBentrySTDinterwordspacing

\bibitem{Firth1957}
J.~Firth, ``A synopsis of linguistic theory 1930-1955,'' in \emph{Studies in
  Linguistic Analysis}.\hskip 1em plus 0.5em minus 0.4em\relax Philological
  Society, Oxford, 1957, reprinted in Palmer, F. (ed. 1968) Selected Papers of
  J. R. Firth, Longman, Harlow.

\bibitem{stahlberg2020neural}
F.~Stahlberg, ``Neural machine translation: A review,'' \emph{Journal of
  Artificial Intelligence Research}, vol.~69, pp. 343--418, 2020.

\bibitem{de2020sign}
M.~De~Coster, M.~Van~Herreweghe, and J.~Dambre, ``Sign language recognition
  with transformer networks,'' in \emph{12th International Conference on
  Language Resources and Evaluation}.\hskip 1em plus 0.5em minus 0.4em\relax
  European Language Resources Association (ELRA), 2020, pp. 6018--6024.

\bibitem{cao2019openpose}
Z.~Cao, G.~Hidalgo, T.~Simon, S.-E. Wei, and Y.~Sheikh, ``Openpose: realtime
  multi-person 2d pose estimation using part affinity fields,'' \emph{IEEE
  transactions on pattern analysis and machine intelligence}, vol.~43, no.~1,
  pp. 172--186, 2019.

\bibitem{camgoz2018neural}
N.~C. Camgoz, S.~Hadfield, O.~Koller, H.~Ney, and R.~Bowden, ``Neural sign
  language translation,'' in \emph{Proceedings of the IEEE Conference on
  Computer Vision and Pattern Recognition}, 2018, pp. 7784--7793.

\bibitem{camgoz2020sign}
N.~C. Camgoz, O.~Koller, S.~Hadfield, and R.~Bowden, ``Sign language
  transformers: Joint end-to-end sign language recognition and translation,''
  in \emph{Proceedings of the IEEE/CVF conference on computer vision and
  pattern recognition}, 2020, pp. 10\,023--10\,033.

\bibitem{yin2020better}
K.~Yin and J.~Read, ``Better sign language translation with stmc-transformer,''
  in \emph{Proceedings of the 28th International Conference on Computational
  Linguistics}, 2020, pp. 5975--5989.

\bibitem{stein2012analysis}
D.~Stein, C.~Schmidt, and H.~Ney, ``Analysis, preparation, and optimization of
  statistical sign language machine translation,'' \emph{Machine Translation},
  vol.~26, no.~4, pp. 325--357, 2012.

\bibitem{morrissey2007combining}
S.~Morrissey, A.~Way, D.~Stein, J.~Bungeroth, and H.~Ney, ``Combining
  data-driven mt systems for improved sign language translation,'' in
  \emph{European Association for Machine Translation}, 2007.

\bibitem{forster2014extensions}
J.~Forster, C.~Schmidt, O.~Koller, M.~Bellgardt, and H.~Ney, ``Extensions of
  the sign language recognition and translation corpus rwth-phoenix-weather.''
  in \emph{LREC}, 2014, pp. 1911--1916.

\bibitem{stein2007hand}
\BIBentryALTinterwordspacing
D.~Stein, P.~Dreuw, H.~Ney, S.~Morrissey, and A.~Way, ``Hand in hand: automatic
  sign language to {E}nglish translation,'' in \emph{Proceedings of the 11th
  Conference on Theoretical and Methodological Issues in Machine Translation of
  Natural Languages: Papers}, Sk{\"o}vde, Sweden, Sep. 7-9 2007. [Online].
  Available: \url{https://aclanthology.org/2007.tmi-papers.26}
\BIBentrySTDinterwordspacing

\bibitem{stein2010sign}
D.~Stein, C.~Schmidt, and H.~Ney, ``Sign language machine translation
  overkill,'' in \emph{International Workshop on Spoken Language Translation
  (IWSLT) 2010}, 2010.

\bibitem{lopez2010spanish}
V.~L{\'o}pez, R.~San-Segundo, R.~Mart{\'\i}n, J.~M. Lucas, and J.~D. Echeverry,
  ``Spanish generation from spanish sign language using a phrase-based
  translation system,'' \emph{technology}, vol.~9, p.~10, 2010.

\bibitem{dreuw2008spoken}
P.~Dreuw, D.~Stein, T.~Deselaers, D.~Rybach, M.~Zahedi, J.~Bungeroth, and
  H.~Ney, ``Spoken language processing techniques for sign language recognition
  and translation,'' \emph{Technology and Disability}, vol.~20, no.~2, pp.
  121--133, 2008.

\bibitem{bungeroth2004statistical}
J.~Bungeroth and H.~Ney, ``Statistical sign language translation,'' in
  \emph{Workshop on representation and processing of sign languages, LREC},
  vol.~4.\hskip 1em plus 0.5em minus 0.4em\relax Citeseer, 2004, pp. 105--108.

\bibitem{schmidt2013using}
C.~Schmidt, O.~Koller, H.~Ney, T.~Hoyoux, and J.~Piater, ``Using viseme
  recognition to improve a sign language translation system,'' in
  \emph{International Workshop on Spoken Language Translation}.\hskip 1em plus
  0.5em minus 0.4em\relax Citeseer, 2013, pp. 197--203.

\bibitem{luqman2020machine}
H.~Luqman and S.~A. Mahmoud, ``A machine translation system from arabic sign
  language to arabic,'' \emph{Universal Access in the Information Society},
  vol.~19, no.~4, pp. 891--904, 2020.

\bibitem{zheng2020improved}
J.~Zheng, Z.~Zhao, M.~Chen, J.~Chen, C.~Wu, Y.~Chen, X.~Shi, and Y.~Tong, ``An
  improved sign language translation model with explainable adaptations for
  processing long sign sentences,'' \emph{Computational Intelligence and
  Neuroscience}, vol. 2020, 2020.

\bibitem{camgoz2020multi}
N.~C. Camgoz, O.~Koller, S.~Hadfield, and R.~Bowden, ``Multi-channel
  transformers for multi-articulatory sign language translation,'' in
  \emph{European Conference on Computer Vision}.\hskip 1em plus 0.5em minus
  0.4em\relax Springer, 2020, pp. 301--319.

\bibitem{orbay2020neural}
A.~Orbay and L.~Akarun, ``Neural sign language translation by learning
  tokenization,'' in \emph{2020 15th IEEE International Conference on Automatic
  Face and Gesture Recognition (FG 2020)}.\hskip 1em plus 0.5em minus
  0.4em\relax IEEE, 2020, pp. 222--228.

\bibitem{zhao2021conditional}
J.~Zhao, W.~Qi, W.~Zhou, D.~Nan, M.~Zhou, and H.~Li, ``Conditional sentence
  generation and cross-modal reranking for sign language translation,''
  \emph{IEEE Transactions on Multimedia}, 2021.

\bibitem{zhou2021improving}
H.~Zhou, W.~Zhou, W.~Qi, J.~Pu, and H.~Li, ``Improving sign language
  translation with monolingual data by sign back-translation,'' in
  \emph{Proceedings of the IEEE/CVF Conference on Computer Vision and Pattern
  Recognition}, 2021, pp. 1316--1325.

\bibitem{zhou2021spatial}
H.~Zhou, W.~Zhou, Y.~Zhou, and H.~Li, ``Spatial-temporal multi-cue network for
  sign language recognition and translation,'' \emph{IEEE Transactions on
  Multimedia}, 2021.

\bibitem{de-coster-etal-2021-frozen}
\BIBentryALTinterwordspacing
M.~De~Coster, K.~D{'}Oosterlinck, M.~Pizurica, P.~Rabaey, S.~Verlinden,
  M.~Van~Herreweghe, and J.~Dambre, ``Frozen pretrained transformers for neural
  sign language translation,'' in \emph{Proceedings of the 1st International
  Workshop on Automatic Translation for Signed and Spoken Languages
  (AT4SSL)}.\hskip 1em plus 0.5em minus 0.4em\relax Virtual: Association for
  Machine Translation in the Americas, Aug. 2021, pp. 88--97. [Online].
  Available: \url{https://aclanthology.org/2021.mtsummit-at4ssl.10}
\BIBentrySTDinterwordspacing

\bibitem{ko2019neural}
S.-K. Ko, C.~J. Kim, H.~Jung, and C.~Cho, ``Neural sign language translation
  based on human keypoint estimation,'' \emph{Applied Sciences}, vol.~9,
  no.~13, p. 2683, 2019.

\bibitem{guo2019hierarchical}
D.~Guo, W.~Zhou, A.~Li, H.~Li, and M.~Wang, ``Hierarchical recurrent deep
  fusion using adaptive clip summarization for sign language translation,''
  \emph{IEEE Transactions on Image Processing}, vol.~29, pp. 1575--1590, 2019.

\bibitem{kim2020robust}
S.~Kim, C.~J. Kim, H.-M. Park, Y.~Jeong, J.~Y. Jang, and H.~Jung, ``Robust
  keypoint normalization method for korean sign language translation using
  transformer,'' in \emph{2020 International Conference on Information and
  Communication Technology Convergence (ICTC)}.\hskip 1em plus 0.5em minus
  0.4em\relax IEEE, 2020, pp. 1303--1305.

\bibitem{guo2018hierarchical}
D.~Guo, W.~Zhou, H.~Li, and M.~Wang, ``Hierarchical lstm for sign language
  translation,'' in \emph{Proceedings of the AAAI Conference on Artificial
  Intelligence}, vol.~32, no.~1, 2018.

\bibitem{kumar2018time}
S.~S. Kumar, T.~Wangyal, V.~Saboo, and R.~Srinath, ``Time series neural
  networks for real time sign language translation,'' in \emph{2018 17th IEEE
  International Conference on Machine Learning and Applications (ICMLA)}.\hskip
  1em plus 0.5em minus 0.4em\relax IEEE, 2018, pp. 243--248.

\bibitem{rodriguez2021important}
J.~Rodriguez and F.~Martinez, ``How important is motion in sign language
  translation?'' \emph{IET Computer Vision}, vol.~15, no.~3, pp. 224--234,
  2021.

\bibitem{arvanitis2019translation}
N.~Arvanitis, C.~Constantinopoulos, and D.~Kosmopoulos, ``Translation of sign
  language glosses to text using sequence-to-sequence attention models,'' in
  \emph{2019 15th International Conference on Signal-Image Technology \&
  Internet-Based Systems (SITIS)}.\hskip 1em plus 0.5em minus 0.4em\relax IEEE,
  2019, pp. 296--302.

\bibitem{rodriguez2020understanding}
J.~Rodriguez, J.~Chacon, E.~Rangel, L.~Guayacan, C.~Hernandez, L.~Hernandez,
  and F.~Martinez, ``Understanding motion in sign language: A new structured
  translation dataset,'' in \emph{Proceedings of the Asian Conference on
  Computer Vision}, 2020.

\bibitem{partaourides2020variational}
H.~Partaourides, A.~Voskou, D.~Kosmopoulos, S.~Chatzis, and D.~N. Metaxas,
  ``Variational bayesian sequence-to-sequence networks for memory-efficient
  sign language translation,'' in \emph{International Symposium on Visual
  Computing}.\hskip 1em plus 0.5em minus 0.4em\relax Springer, 2020, pp.
  251--262.

\bibitem{moryossef-etal-2021-data}
\BIBentryALTinterwordspacing
A.~Moryossef, K.~Yin, G.~Neubig, and Y.~Goldberg, ``Data augmentation for sign
  language gloss translation,'' in \emph{Proceedings of the 1st International
  Workshop on Automatic Translation for Signed and Spoken Languages
  (AT4SSL)}.\hskip 1em plus 0.5em minus 0.4em\relax Virtual: Association for
  Machine Translation in the Americas, Aug. 2021, pp. 1--11. [Online].
  Available: \url{https://aclanthology.org/2021.mtsummit-at4ssl.1}
\BIBentrySTDinterwordspacing

\bibitem{moe2020unsupervised}
S.~Z. Moe, Y.~K. Thu, H.~A. Thant, N.~W. Min, and T.~Supnithi, ``Unsupervised
  neural machine translation between myanmar sign language and myanmar
  language,'' \emph{tic}, vol.~14, no.~15, p.~16, 2020.

\bibitem{zhang-duh-2021-approaching}
\BIBentryALTinterwordspacing
X.~Zhang and K.~Duh, ``Approaching sign language gloss translation as a
  low-resource machine translation task,'' in \emph{Proceedings of the 1st
  International Workshop on Automatic Translation for Signed and Spoken
  Languages (AT4SSL)}.\hskip 1em plus 0.5em minus 0.4em\relax Virtual:
  Association for Machine Translation in the Americas, Aug. 2021, pp. 60--70.
  [Online]. Available: \url{https://aclanthology.org/2021.mtsummit-at4ssl.7}
\BIBentrySTDinterwordspacing

\bibitem{luong-etal-2015-effective}
\BIBentryALTinterwordspacing
T.~Luong, H.~Pham, and C.~D. Manning, ``Effective approaches to attention-based
  neural machine translation,'' in \emph{Proceedings of the 2015 Conference on
  Empirical Methods in Natural Language Processing}.\hskip 1em plus 0.5em minus
  0.4em\relax Lisbon, Portugal: Association for Computational Linguistics, Sep.
  2015, pp. 1412--1421. [Online]. Available:
  \url{https://aclanthology.org/D15-1166}
\BIBentrySTDinterwordspacing

\bibitem{deng2009imagenet}
J.~Deng, W.~Dong, R.~Socher, L.-J. Li, K.~Li, and L.~Fei-Fei, ``Imagenet: A
  large-scale hierarchical image database,'' in \emph{2009 IEEE conference on
  computer vision and pattern recognition}.\hskip 1em plus 0.5em minus
  0.4em\relax Ieee, 2009, pp. 248--255.

\bibitem{forster2012rwth}
J.~Forster, C.~Schmidt, T.~Hoyoux, O.~Koller, U.~Zelle, J.~H. Piater, and
  H.~Ney, ``Rwth-phoenix-weather: A large vocabulary sign language recognition
  and translation corpus.'' in \emph{LREC}, vol.~9, 2012, pp. 3785--3789.

\bibitem{othman2012english}
A.~Othman and M.~Jemni, ``English-asl gloss parallel corpus 2012: Aslg-pc12,''
  in \emph{5th Workshop on the Representation and Processing of Sign Languages:
  Interactions between Corpus and Lexicon LREC}, 2012.

\bibitem{EsplGomis2019ParaCrawlWP}
M.~Espl{\`a}-Gomis, M.~Forcada, G.~Ram{\'i}rez-S{\'a}nchez, and H.~T. Hoang,
  ``Paracrawl: Web-scale parallel corpora for the languages of the eu,'' in
  \emph{MTSummit}, 2019.

\bibitem{krizhevsky2012imagenet}
A.~Krizhevsky, I.~Sutskever, and G.~E. Hinton, ``Imagenet classification with
  deep convolutional neural networks,'' \emph{Advances in neural information
  processing systems}, vol.~25, pp. 1097--1105, 2012.

\bibitem{moryossef2021evaluating}
A.~Moryossef, I.~Tsochantaridis, J.~Dinn, N.~C. Camgoz, R.~Bowden, T.~Jiang,
  A.~Rios, M.~Muller, and S.~Ebling, ``Evaluating the immediate applicability
  of pose estimation for sign language recognition,'' in \emph{Proceedings of
  the IEEE/CVF Conference on Computer Vision and Pattern Recognition}, 2021,
  pp. 3434--3440.

\bibitem{Wolf_Transformers_StateoftheArt_Natural_2020}
\BIBentryALTinterwordspacing
T.~Wolf, L.~Debut, V.~Sanh, J.~Chaumond, C.~Delangue, A.~Moi, P.~Cistac, C.~Ma,
  Y.~Jernite, J.~Plu, C.~Xu, T.~Le~Scao, S.~Gugger, M.~Drame, Q.~Lhoest, and
  A.~M. Rush, ``{Transformers: State-of-the-Art Natural Language Processing},''
  pp. 38--45, 10 2020. [Online]. Available:
  \url{https://www.aclweb.org/anthology/2020.emnlp-demos.6}
\BIBentrySTDinterwordspacing

\bibitem{de2021isolated}
M.~De~Coster, M.~Van~Herreweghe, and J.~Dambre, ``Isolated sign recognition
  from rgb video using pose flow and self-attention,'' in \emph{Proceedings of
  the IEEE/CVF Conference on Computer Vision and Pattern Recognition}, 2021,
  pp. 3441--3450.

\bibitem{callison2006re}
C.~Callison-Burch, M.~Osborne, and P.~Koehn, ``Re-evaluating the role of {BLEU}
  in machine translation research,'' in \emph{11th Conference of the European
  Chapter of the Association for Computational Linguistics}, 2006.

\bibitem{chen2020simple}
T.~Chen, S.~Kornblith, M.~Norouzi, and G.~Hinton, ``A simple framework for
  contrastive learning of visual representations,'' in \emph{International
  conference on machine learning}.\hskip 1em plus 0.5em minus 0.4em\relax PMLR,
  2020, pp. 1597--1607.

\bibitem{grill2020bootstrap}
J.-B. Grill, F.~Strub, F.~Altch{\'e}, C.~Tallec, P.~Richemond, E.~Buchatskaya,
  C.~Doersch, B.~Pires, Z.~Guo, M.~Azar \emph{et~al.}, ``Bootstrap your own
  latent: A new approach to self-supervised learning,'' in \emph{Neural
  Information Processing Systems}, 2020.

\bibitem{caron2021emerging}
M.~Caron, H.~Touvron, I.~Misra, H.~J\'egou, J.~Mairal, P.~Bojanowski, and
  A.~Joulin, ``Emerging properties in self-supervised vision transformers,'' in
  \emph{Proceedings of the International Conference on Computer Vision (ICCV)},
  2021.

\bibitem{baevski2020wav2vec}
A.~Baevski, Y.~Zhou, A.~Mohamed, and M.~Auli, ``wav2vec 2.0: A framework for
  self-supervised learning of speech representations,'' \emph{Advances in
  Neural Information Processing Systems}, vol.~33, 2020.

\bibitem{de-meulder-2021-good}
\BIBentryALTinterwordspacing
M.~De~Meulder, ``Is {``}good enough{''} good enough? ethical and responsible
  development of sign language technologies,'' in \emph{Proceedings of the 1st
  International Workshop on Automatic Translation for Signed and Spoken
  Languages (AT4SSL)}.\hskip 1em plus 0.5em minus 0.4em\relax Virtual:
  Association for Machine Translation in the Americas, Aug. 2021, pp. 12--22.
  [Online]. Available: \url{https://aclanthology.org/2021.mtsummit-at4ssl.2}
\BIBentrySTDinterwordspacing

\bibitem{vanmassenhove-etal-2019-lost}
\BIBentryALTinterwordspacing
E.~Vanmassenhove, D.~Shterionov, and A.~Way, ``Lost in translation: Loss and
  decay of linguistic richness in machine translation,'' in \emph{Proceedings
  of Machine Translation Summit XVII Volume 1: Research Track}.\hskip 1em plus
  0.5em minus 0.4em\relax Dublin, Ireland: European Association for Machine
  Translation, Aug. 2019, pp. 222--232. [Online]. Available:
  \url{https://www.aclweb.org/anthology/W19-6622}
\BIBentrySTDinterwordspacing

\bibitem{vanmassenhove-2021-MachineTranslationese}
\BIBentryALTinterwordspacing
E.~Vanmassenhove, D.~Shterionov, and M.~Gwilliam, ``Machine translationese:
  Effects of algorithmic bias on linguistic complexity in machine
  translation,'' in \emph{Proceedings of the 16th Conference of the European
  Chapter of the Association for Computational Linguistics: Main Volume, {EACL}
  2021, Online, April 19 - 23, 2021}, P.~Merlo, J.~Tiedemann, and R.~Tsarfaty,
  Eds.\hskip 1em plus 0.5em minus 0.4em\relax Association for Computational
  Linguistics, 2021, pp. 2203--2213. [Online]. Available:
  \url{https://aclanthology.org/2021.eacl-main.188/}
\BIBentrySTDinterwordspacing

\bibitem{bragg2019sign}
D.~Bragg, O.~Koller, M.~Bellard, L.~Berke, P.~Boudreault, A.~Braffort,
  N.~Caselli, M.~Huenerfauth, H.~Kacorri, T.~Verhoef \emph{et~al.}, ``Sign
  language recognition, generation, and translation: An interdisciplinary
  perspective,'' in \emph{The 21st international ACM SIGACCESS conference on
  computers and accessibility}, 2019, pp. 16--31.

\bibitem{sincan2020autsl}
O.~M. Sincan and H.~Y. Keles, ``Autsl: A large scale multi-modal turkish sign
  language dataset and baseline methods,'' \emph{IEEE Access}, vol.~8, pp.
  181\,340--181\,355, 2020.

\bibitem{gholami2021survey}
A.~Gholami, S.~Kim, Z.~Dong, Z.~Yao, M.~W. Mahoney, and K.~Keutzer, ``A survey
  of quantization methods for efficient neural network inference,'' \emph{arXiv
  preprint arXiv:2103.13630}, 2021.

\bibitem{wang2020linformer}
S.~Wang, B.~Z. Li, M.~Khabsa, H.~Fang, and H.~Ma, ``Linformer: Self-attention
  with linear complexity,'' \emph{arXiv preprint arXiv:2006.04768}, 2020.

\bibitem{tan2019efficientnet}
M.~Tan and Q.~Le, ``Efficientnet: Rethinking model scaling for convolutional
  neural networks,'' in \emph{International Conference on Machine
  Learning}.\hskip 1em plus 0.5em minus 0.4em\relax PMLR, 2019, pp. 6105--6114.

\bibitem{strubell-etal-2019-energy}
\BIBentryALTinterwordspacing
E.~Strubell, A.~Ganesh, and A.~McCallum, ``Energy and policy considerations for
  deep learning in {NLP},'' in \emph{Proceedings of the 57th Annual Meeting of
  the Association for Computational Linguistics}.\hskip 1em plus 0.5em minus
  0.4em\relax Florence, Italy: Association for Computational Linguistics, Jul.
  2019, pp. 3645--3650. [Online]. Available:
  \url{https://www.aclweb.org/anthology/P19-1355}
\BIBentrySTDinterwordspacing

\bibitem{Schwartz2020GreenA}
R.~Schwartz, J.~Dodge, N.~A. Smith, and O.~Etzioni, ``Green ai,''
  \emph{Communications of the ACM}, vol.~63, pp. 54 -- 63, 2020.

\bibitem{sennrich-zhang-2019-revisiting}
\BIBentryALTinterwordspacing
R.~Sennrich and B.~Zhang, ``Revisiting low-resource neural machine translation:
  A case study,'' in \emph{Proceedings of the 57th Annual Meeting of the
  Association for Computational Linguistics}.\hskip 1em plus 0.5em minus
  0.4em\relax Florence, Italy: Association for Computational Linguistics, Jul.
  2019, pp. 211--221. [Online]. Available:
  \url{https://aclanthology.org/P19-1021}
\BIBentrySTDinterwordspacing

\bibitem{wolfe2021myth}
R.~Wolfe, J.~Mcdonald, E.~Efthimiou, E.~Fotinea, F.~Picron, D.~Van~Landuyt,
  T.~Sioen, A.~Braffort, M.~Filhol, S.~Ebling \emph{et~al.}, ``The myth of
  signing avatars,'' in \emph{1st International Workshop on Automatic
  Translation for Signed and Spoken Languages}, 2021.

\bibitem{shterionov2021signon}
D.~Shterionov, V.~Vandeghinste, H.~Saggion, J.~Blat, M.~De~Coster, J.~Dambre,
  H.~van~den Heuvel, I.~Murtagh, L.~Leeson, and I.~Schuurman, ``{The SignON
  project: a Sign Language Translation Framework},'' in \emph{31st Meeting of
  Computational Linguists in The Netherlands (CLIN)}, 2021.

\bibitem{easier}
\BIBentryALTinterwordspacing
E.~PROJECT, ``Easier -- intelligent automatic sign language translation,''
  2021. [Online]. Available: \url{https://www.project-easier.eu/}
\BIBentrySTDinterwordspacing

\end{thebibliography}

\end{document}